\documentclass{article}

\usepackage{microtype}
\usepackage{graphicx}
\usepackage{subcaption}
\usepackage{booktabs}

\usepackage{hyperref}

\usepackage[preprint]{colm2026_conference}

\usepackage[T1]{fontenc}
\usepackage{amsmath}
\usepackage{amssymb}
\usepackage{amsthm}
\usepackage{url}
\usepackage{enumitem}
\usepackage{listings}
\usepackage{xcolor}
\usepackage[most]{tcolorbox}
\usepackage{xspace}

\newtheorem{definition}{Definition}

\definecolor{darkblue}{rgb}{0, 0, 0.5}
\hypersetup{colorlinks=true, citecolor=darkblue, linkcolor=darkblue, urlcolor=darkblue}

\title{TRACE: Capability-Targeted Agentic Training}

\author{Hangoo Kang\thanks{Equal contribution.} \quad Tarun Suresh\footnotemark[1] \quad
Jon Saad-Falcon \quad Azalia Mirhoseini \\
Stanford University \\
\texttt{\{hangook, tsuresh\}@stanford.edu}}

\begin{document}

\newcommand{\tbench}{$\tau^2$-Bench\xspace}
\newcommand{\tsb}{ToolSandBox\xspace}
\newcommand{\swe}{SWE-bench Verified\xspace}
\newcommand{\tool}{TRACE\xspace}

\maketitle

\begin{abstract}
Models fail to complete agentic tasks often due to missing core capabilities. However, mainstream approaches for addressing these failures often rely on fine-tuning directly on the target environments or generating synthetic data that is not targeted to the model's actual capability deficits, resulting in low sample efficiency and limited generalizability. We introduce \tool{} (Turning Recurrent Agent failures into Capability-targeted training Environments), an end-to-end system for environment-specific agent self-improvement. \tool{} contrasts successful and failed trajectories to automatically identify lacking capabilities, synthesizes a targeted training environment for each that rewards whether the capability was exercised, trains a LoRA adapter via RL on each synthetic environment, and then trains a mixture-of-experts (MoE) model over the capability adapters. \tool{} can be effectively applied across different environments, improving over the base agent by \textbf{+15.3 points} on \tbench{} (customer service) and \textbf{+15 points} Pass@1 on \swe{} (software engineering), outperforming the strongest external baselines, GEPA and SWE-RL, by \textbf{+8.6 points} and \textbf{+8.4 points}, respectively. In addition, \tool{} is more sample-efficient than strong finetuning baselines: using less than one-fourth the number of rollouts, \tool{} outperforms the best-performing baselines, GRPO and GEPA, and achieves higher final accuracy by \textbf{+10.4} and \textbf{+8.6} points on \tbench{}. The code is available at \url{https://github.com/ScalingIntelligence/TRACE.git}.
\end{abstract}

\section{Introduction}

\begin{figure*}[h]
\centering
\begin{subfigure}{0.49\textwidth}
\centering
\includegraphics[width=\textwidth]{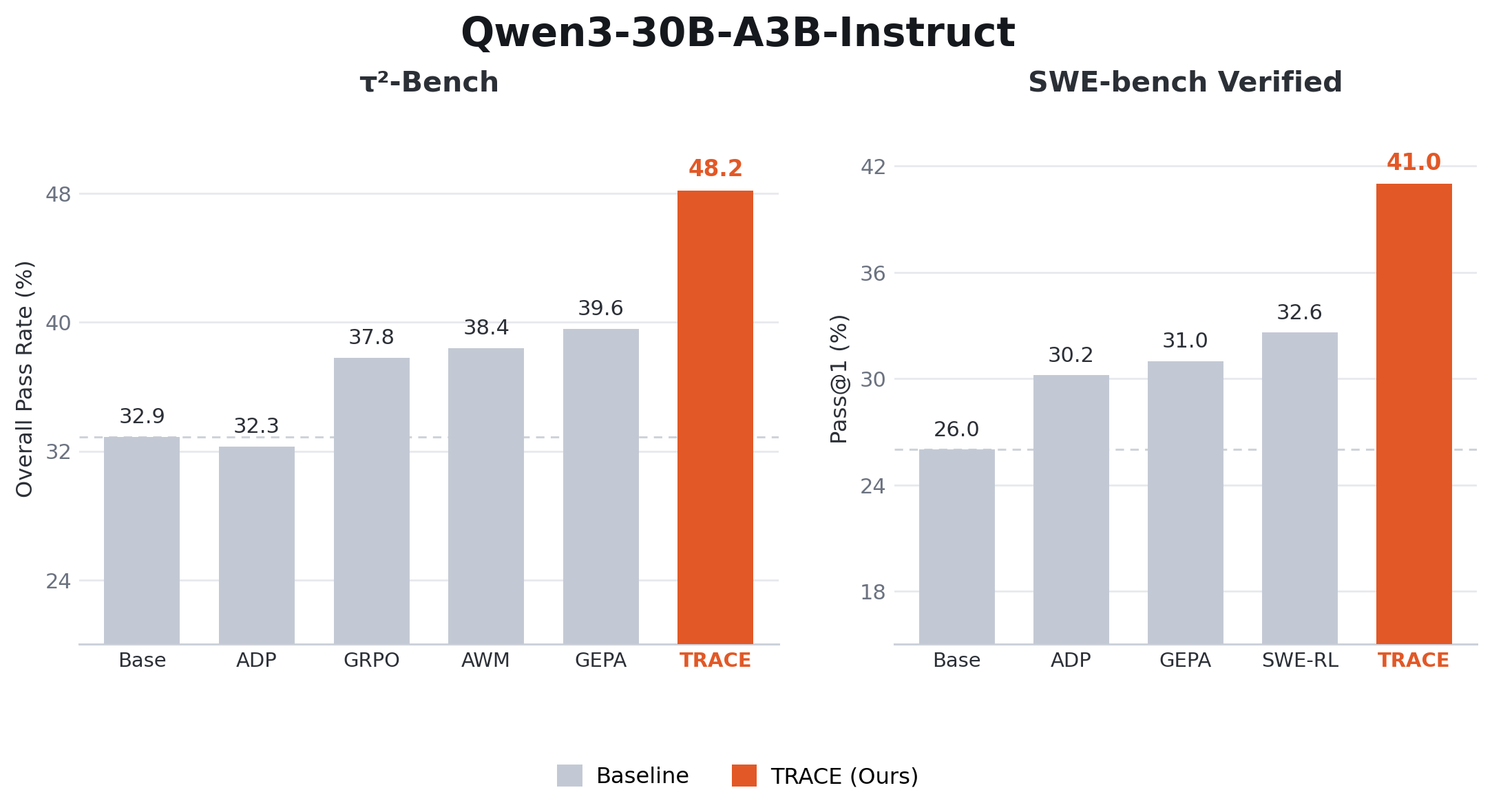}
\caption{Qwen3-30B-A3B-Instruct}
\label{fig:teaser_30b}
\end{subfigure}
\hfill
\begin{subfigure}{0.49\textwidth}
\centering
\includegraphics[width=\textwidth]{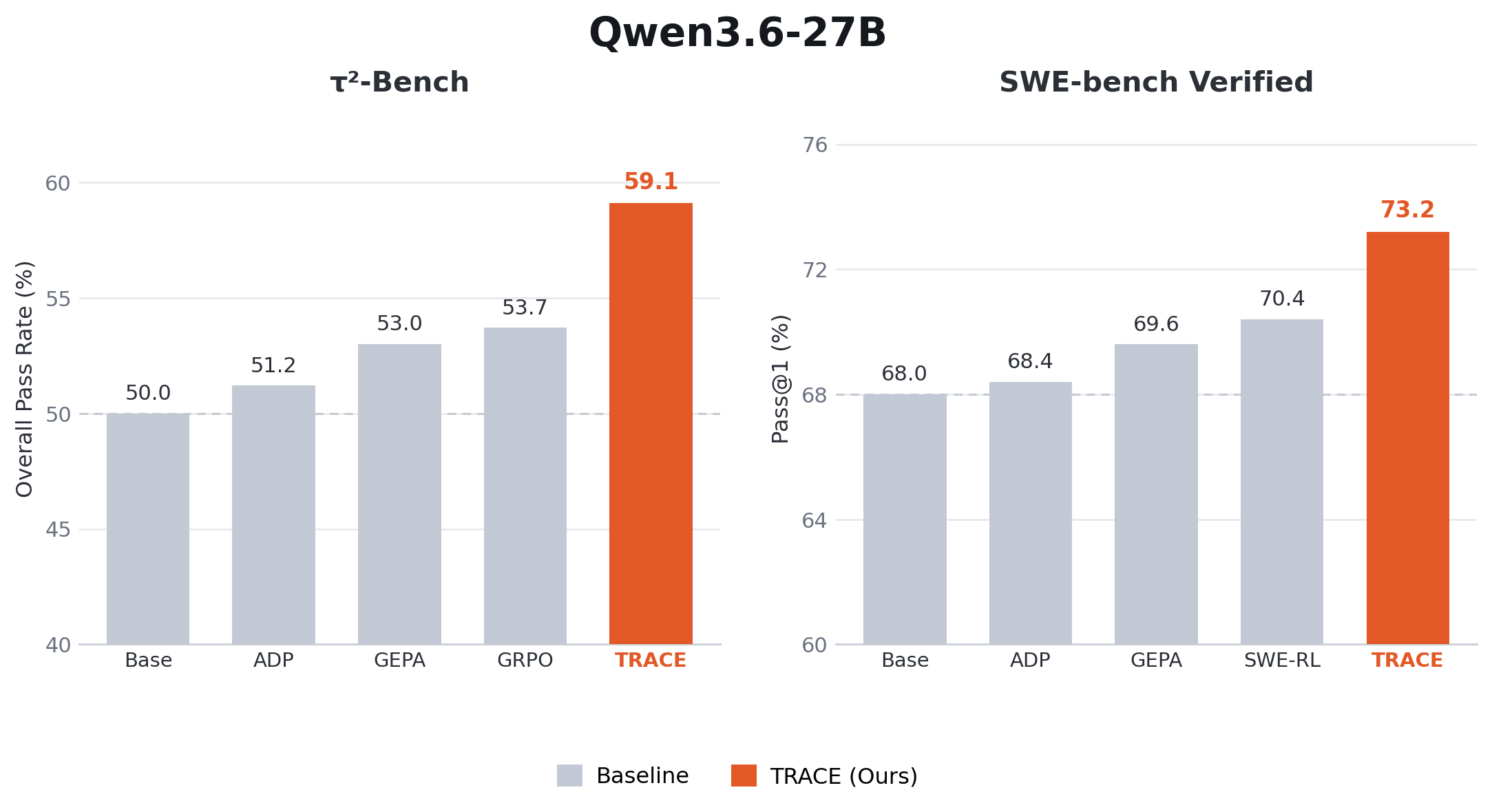}
\caption{Qwen3.6-27B}
\label{fig:teaser_27b}
\end{subfigure}
\caption{\tool{} outperforms the base agent and all baselines on both \tbench{} (overall pass rate) and \swe{} Verified (Pass@1), across two agent backbones.}
\label{fig:teaser}
\end{figure*}

Large Language Models (LLMs) are increasingly deployed in agentic environments, such as customer-service workflows~\citep{yao2024taubenchbenchmarktoolagentuserinteraction} and coding platforms~\citep{jimenez2024swebenchlanguagemodelsresolve, xu2025theagentcompanybenchmarkingllmagents}, where they must complete multi-step tasks through tool use, reasoning, and interaction with an environment. When models fail in these settings, their failures are often not isolated to a single task instance. Instead, they frequently reflect missing core capabilities that are required across many different tasks. We define a \emph{capability} as performing one or more actions in a trajectory that are necessary for successfully solving some subset of task instances in the target environment. For example, in a customer-service environment, retrieving the correct customer record is a capability necessary for tasks such as canceling a flight, changing a seat, or verifying a reservation.

A natural approach to improving an LLM in a target agentic environment is to finetune it directly on that environment using Reinforcement Learning (RL) or Supervised Fine-Tuning (SFT). However, direct training on the target environment often provides supervision only at the level of task outcomes or task-specific trajectories. This training signal does not explicitly reveal which underlying capabilities the agent lacks. As a result, the model must implicitly infer shared capabilities from sparse and heterogeneous task-level feedback, making learning sample-inefficient and limiting generalization across task instances.

Another line of work scales synthetic RL environments and training data~\citep{sullivan2025proceduralenvironmentgenerationtooluse, wang2026agentworldmodel, fang2025generalagenticintelligenceenvironment, tu2026scaleenvscalingenvironmentsynthesis, song2026envscalerscalingtoolinteractiveenvironments}. While these approaches increase the amount and diversity of training data, the generated data is often not targeted to the model's actual capability deficits in the target environment. Consequently, training may spend substantial compute on behaviors the model already performs well or on capabilities that are irrelevant to the target failures, yielding limited sample efficiency and weak environment-specific improvement.

\begin{figure*}[t]
\small
\centering
\includegraphics[width=0.92\textwidth]{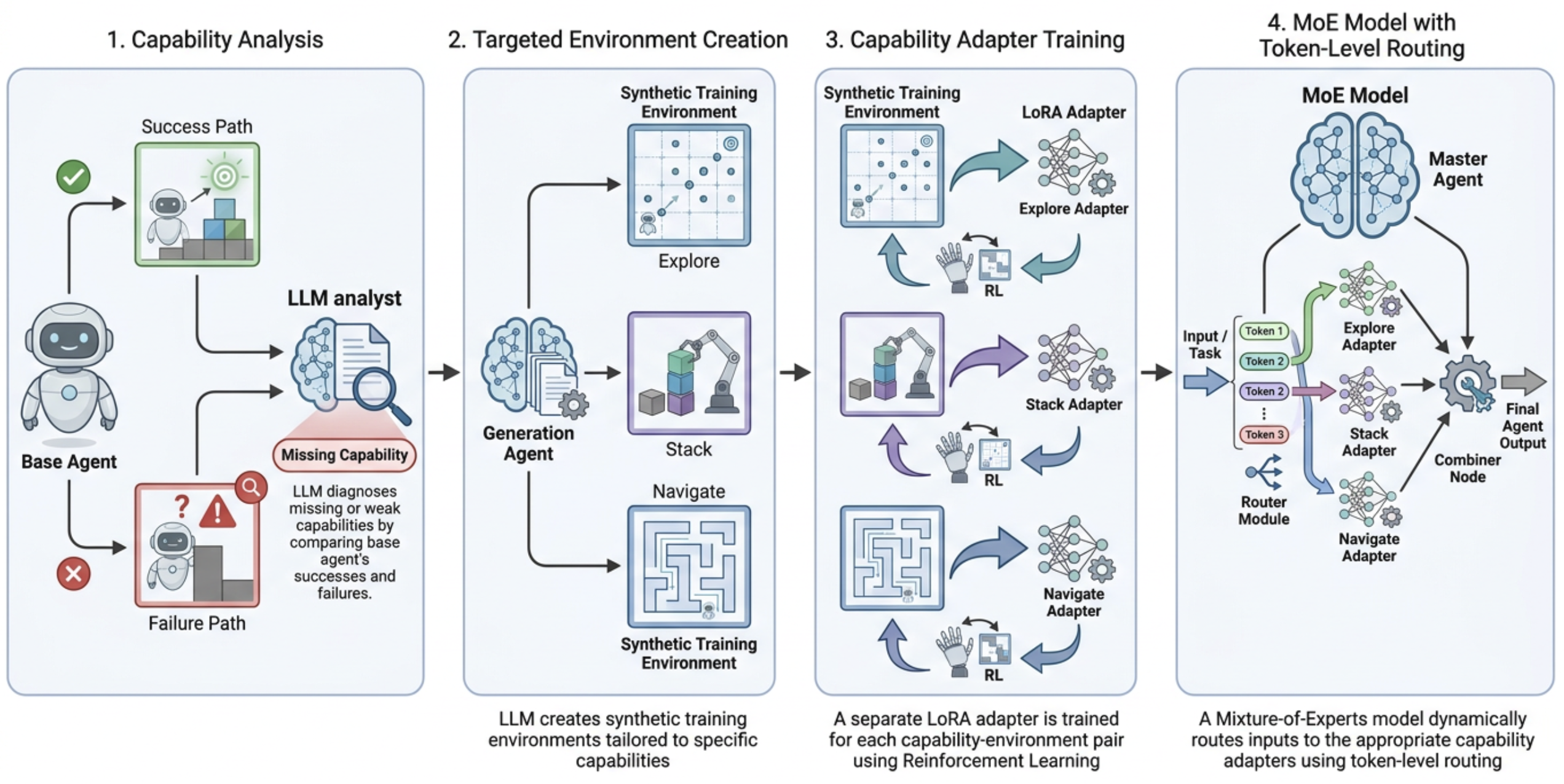}
\caption{Overview of \tool{}, an end-to-end system for automated environment-specific agent self-improvement. \tool{} automatically identifies the specific capabilities that an agent lacks and synthesizes targeted training environments for each capability.}
\label{fig:framework_overview}
\end{figure*}

In this work, we introduce \tool{} (Turning Recurrent Agent failures into Capability-targeted training Environments), an end-to-end system for environment-specific agent self-improvement. Rather than training directly on the target environment or relying on untargeted synthetic data, \tool{} uses the agent's own successful and failed trajectories to identify the capabilities whose absence explains recurring failures. It then synthesizes capability-targeted training environments that isolate and reward exercising each missing capability, trains a LoRA adapter via RL on each synthetic environment, and composes the resulting adapters through a Mixture-of-Experts (MoE) model.

\tool{} is designed around three objectives:

\vspace{-1mm}
\begin{enumerate}[itemsep=0.00pt,topsep=0pt,leftmargin=12pt]
\item \textbf{Identifying high-impact capability deficits.} The agent may fail for many reasons, so it is important to automatically determine which missing capabilities most clearly distinguish failed trajectories from successful ones and prioritize those that account for the largest number of failures.

\item \textbf{Synthesizing capability-targeted training environments.} For each identified deficit, the agent needs a training environment that directly isolates and rewards exercising the missing capability while preserving the target environment's interface.

\item \textbf{Learning and composing multiple capabilities.} The agent must acquire multiple identified capabilities from their respective synthetic environments and apply the appropriate capabilities to each task instance in the target environment.

\end{enumerate}

Figure~\ref{fig:framework_overview} shows an overview of \tool{}, which consists of four steps. In step 1, the base agent generates rollouts in the target environment, and an LLM-based analysis agent contrasts successful and failed trajectories to identify capabilities that meaningfully distinguish the two, ranking them by failure coverage. In step 2, for each retained capability, an LLM-based generation agent uses the capability description and corresponding failed trajectories to construct a synthetic environment that isolates the missing capability while preserving the target environment's interface, such as its tool schemas, interaction protocol, and output format. Task instances are procedurally generated from random seeds, and whether the capability is exercised is verified automatically from tool arguments, state changes, or final outputs. In step 3, we train a LoRA adapter on each capability-specific synthetic environment using RL. In step 4, we train a Mixture-of-Experts (MoE)~\citep{shazeer2017outrageouslylargeneuralnetworks} model over the capability adapters.

Our contributions are as follows:

\begin{itemize}[itemsep=0.00pt,topsep=0pt,leftmargin=12pt]
\item We propose \tool{}, an end-to-end system for environment-specific agent self-improvement. \tool{} contrasts successful and failed trajectories from the target environment to automatically identify the specific capabilities the agent lacks, then synthesizes targeted training environments that teach the agent to exercise those capabilities.

\item We show that \tool{} can be effectively applied across different agentic environments. On \tbench{}~\citep{barres2025tau2benchevaluatingconversationalagents}, a customer-service benchmark, \tool{} achieves a \textbf{48.2\% overall pass rate}, improving over the base agent by \textbf{+15.3 points} and outperforming the strongest external baseline, GEPA, by \textbf{+8.6 points} (Table~\ref{tab:main_tb_results}). On \swe{}~\citep{jimenez2024swebenchlanguagemodelsresolve}, a software-engineering benchmark, \tool{} reaches \textbf{41\% Pass@1}, improving over the base agent by \textbf{+15 points} and outperforming the strongest external baseline, SWE-RL, by \textbf{+8.4 points} (Table~\ref{tab:swe_results}). Moreover, with Qwen3.6-27B, \tool{} attains \textbf{73.2\% Pass@1} on \swe{} Verified, outperforming much larger frontier models, including \textbf{GPT-5.2-Codex} (72.8\%), \textbf{GLM 5}, and \textbf{Claude 4.5 Sonnet}, on the public SWE-bench leaderboard\footnote{\url{https://www.swebench.com/}}, despite using a substantially smaller open-weight model.

\item We demonstrate that \tool{} is more sample-efficient than strong finetuning baselines. With Qwen3-30B-A3B-Instruct, \tool{} uses only 2--4 capability-specific LoRA adapters, each updating only 5.3\% of the model parameters, yet improves performance on \tbench{} by \textbf{+15.3 points} and Pass@1 on \swe{} by \textbf{+15 points}. In addition, using less than one-fourth the number of rollouts, \tool{} outperforms the best-performing baselines, GRPO~\citep{shao2024deepseekmathpushinglimitsmathematical} and GEPA~\citep{agrawal2026gepareflectivepromptevolution}, and achieves higher final accuracy by \textbf{+10.4} and \textbf{+8.6} points on \tbench{}, respectively.

\item We show that training on capability-targeted environments outperforms only adding capabilities to the prompt. With Qwen3-30B-A3B-Instruct, training on capability-targeted environments yields a \textbf{+8.6 point} improvement on \tbench{} and \textbf{+10 point} improvement in Pass@1 on \swe{} over prompt-based optimization (Figure~\ref{fig:cap_scaling}).
\end{itemize}

\section{Related Work}
\label{sec:related_works}
\vspace{-1mm}
\noindent\textbf{LLM Agents and Interactive Environments.}
Autonomous LLM agents are increasingly deployed in complex, multi-turn environments. Rigorous evaluation requires mastering specific tool interfaces and interaction protocols, driving the adoption of benchmarks like \tbench{}, SWE-bench~\citep{jimenez2024swebenchlanguagemodelsresolve}, WebArena~\citep{zhou2024webarenarealisticwebenvironment}, WorkArena~\citep{drouin2024workarenacapablewebagents}, WorkArena++~\citep{boisvert2024workarenacompositionalplanningreasoning}, Terminal-Bench~\citep{merrill2026terminalbenchbenchmarkingagentshard} and TheAgentCompany~\citep{xu2024theagentcompanybenchmarkingllmagents}.

\vspace{-0.5mm}
\noindent\textbf{Agentic Reinforcement Learning and Synthetic Data.}
Recent alignment methods utilize in-the-wild device control~\citep{bai2024digirltraininginthewilddevicecontrol}, implicit step rewards~\citep{liu2025agenticreinforcementlearningimplicit}, or verifiable multi-turn RL~\citep{gao2026fromselfevolvingsyntheticdata, zhang2025agentrlscalingagenticreinforcement}. A common capability-acquisition strategy scales training through procedurally synthesized environments~\citep{sullivan2025proceduralenvironmentgenerationtooluse} (e.g., AWM~\citep{wang2026agentworldmodelinfinity}, EnvScaler~\citep{song2026envscalerscalingtoolinteractiveenvironments}, ScaleEnv~\citep{tu2026scaleenvscalingenvironmentsynthesis}) or unified public trajectories(e.g. ADP~\citep{song2026agentdataprotocolunifyingdatasets}). While effective for general capabilities~\citep{yang2025swesmithscalingdatasoftware, fang2025towardsgeneralagenticintelligence}, these approaches overlook model-specific failures. \tool{} diverges by using contrastive analysis to identify missing capabilities, synthesizing training environments to isolate and reward correct execution of each capability.

\noindent\textbf{LoRA Merging and Routing.}
Recent work on model merging has established general techniques for composing task-specific adaptations~\citep{ilharco2023editingmodelstaskarithmetic, yadav2023tiesmergingresolvinginterference, panariello2025accurateefficientlowrank}. A complementary direction avoids collapsing everything into one checkpoint and instead uses mixtures of LoRA experts, where routing dynamically selects or softly combines experts such as ~\citep{luo2024moeloracontrastivelearningguided, cao2026comolefficientmixturelora}, including foundational LoRA-MoE methods such as LoRAMoE~\citep{dou2024loramoerevolutionizingmixture}, MoLE~\citep{wu2024mixtureofloraexperts}, and MoV/MoLORA~\citep{zadouri2024pushingmixtureexpertslimit}, as well as variants exploring different routing granularities and mergeability~\citep{gao2024higherlayersneedmore, wang2025eachrankcouldbeanexpert}. 
Unlike methods where experts emerge implicitly from multi-task data, \tool{}'s adapters are independently trained on synthetic environments that each isolate a single, contrastively identified capability, yielding an interpretable set of experts. Advances in adapter architectures and MoE training schemes are orthogonal to \tool{} and can be incorporated into our pipeline for further gains.

\section{Method}

\subsection{Preliminaries}

\paragraph{Agentic environment.}
We represent an agentic environment as a tuple $\mathcal{E} = (\mathcal{X}_{\mathcal{E}}, P_{\mathcal{E}}, R_{\mathcal{E}}, y_{\mathcal{E}})$, where $\mathcal{X}_{\mathcal{E}}$ is the set of task instances, $P_{\mathcal{E}}$ defines the interaction dynamics, $R_{\mathcal{E}}(x,\tau)$ is a trajectory-level reward and can either be discrete or continuous, and $y_{\mathcal{E}}(x,\tau) \in \{0,1\}$ is a binary label indicating whether $\tau$ succeeded or not. 

A task instance $x \sim \mathcal{X}_{\mathcal{E}}$ determines the initial observation $o_1$ presented to the agent. An LLM policy $\pi_\theta$ then selects actions conditioned on the interaction history $h_t = (o_1, a_1, \dots, o_t)$ via $a_t \sim \pi_\theta(\cdot \mid h_t)$, where each action is a variable-length token sequence that may include natural-language reasoning and tool calls. The environment returns an observation $o_{t+1}$ after each action, and the episode continues for at most $T$ steps, producing a trajectory $\tau = (o_1, a_1, \dots, o_T, a_T, o_{T+1})$. At episode end, the environment assigns reward $R_{\mathcal{E}}(x,\tau)$ and success label $y_{\mathcal{E}}(x,\tau)$. We denote by $\mathcal{D} = \{(x_i,\tau_i,r_i,y_i)\}_{i=1}^N$ a dataset of $N$ collected episodes.

\vspace{-2mm}
\paragraph{Synthetic environment.}
A synthetic verifiable environment is an agentic environment $\mathcal{E}_s = (G_{\mathcal{E}_s}, P_{\mathcal{E}_s}, R_{\mathcal{E}_s}, y_{\mathcal{E}_s})$ whose task instances are produced by an explicit generator $G_{\mathcal{E}_s}$ and whose reward and success label can be evaluated automatically from the task instance and trajectory. Concretely, given a random seed $z$, the generator produces a task instance $x_s = G_{\mathcal{E}_s}(z)$ together with the associated environment configuration, such as the initial state, transition logic, and evaluation criteria. Interaction with $\mathcal{E}_s$ then produces a trajectory $\tau_s$ with reward $R_{\mathcal{E}_s}(x_s,\tau_s)$ and success label $y_{\mathcal{E}_s}(x_s,\tau_s)$. Because both generation and verification are algorithmic, synthetic verifiable environments can be scaled to produce large numbers of training instances with verifiable reward signal.

\vspace{-0.5mm}
\subsection{Problem Formulation}

Our goal is to learn an LLM policy $\pi_\theta$ for a target agentic environment $\mathcal{E}$ that maximizes expected return:
\[
J(\pi_\theta;\mathcal{E})
\;=\;
\mathbb{E}_{x \sim \mathcal{X}_{\mathcal{E}},\, \tau \sim p_{\pi_\theta}(\cdot \mid x,\mathcal{E})}
\bigl[ R_{\mathcal{E}}(x,\tau) \bigr],
\]
Here, $x$ is a task instance, $\tau$ is a corresponding trajectory induced by executing $\pi_\theta$ in $\mathcal{E}$ on $x$, and $R_{\mathcal{E}}(x,\tau)$ is the trajectory-level reward.

A central challenge is that optimization in the target environment provides limited failure attribution. The policy must infer what is necessary for success across task instances only indirectly from the training signal, making optimization inefficient. We formalize such underlying necessities as \emph{capabilities}.

\begin{definition}[Capability]
A capability consists of performing one or more actions in a trajectory that are necessary for successfully solving some subset of task instances in the environment. For $X_c \subseteq X_\mathcal{E}$, let $\nu_c(x,\tau)=1$ indicate that capability $c$ is exercised in trajectory $\tau$ for task instance $x$. Then, for every $x \in X_c$ and a corresponding trajectory $\tau$,
$
y_\mathcal{E}(x,\tau)=1 \;\Rightarrow\; \nu_c(x,\tau)=1.
$
\end{definition}

\noindent\textbf{Examples.}
\begin{enumerate}
    \item In a customer-service environment, retrieving the correct customer record is a capability, since it is necessary for tasks such as canceling a flight, changing a seat, or verifying a reservation.
    \item In a coding environment, correctly locating the relevant function or file is a capability, since it is necessary for tasks such as fixing a bug, adding a feature, or updating an API call.
\end{enumerate}

However, in $\mathcal{E}$, $R_{\mathcal{E}}(x,\tau)$ does not generally provide an explicit signal of whether $\nu_c(x,\tau)=1$ for each capability $c$, making it difficult to isolate and train any single capability directly. To address this, we introduce:

\begin{definition}[Capability-targeted synthetic environment]
For a capability $c$, a capability-targeted synthetic environment is a synthetic verifiable environment $\mathcal{E}_s^c = (G_c, P_c, R_c, y_c)$ constructed to satisfy three properties:
\begin{enumerate}
    \item Every task instance $x_c$ generated by $G_c$ ensures that exercising $c$ is necessary for success.
    \item $P_c$ preserves the aspects of $P_{\mathcal{E}}$ that are relevant to exercising $c$, such as tool schemas, state representation, and policy constraints.
    \item For any $x_c$ and a corresponding trajectory $\tau_c$, the reward $R_c(x_c,\tau_c)$ and success label $y_c(x_c,\tau_c)$ are automatically computable from $(x_c,\tau_c)$. Higher reward and success depend primarily on whether $c$ is exercised in $\tau_c$.
\end{enumerate}
\end{definition}

Because each $\mathcal{E}_s^c$ isolates a single capability, its reward signal is generally denser and more attributable than that of $\mathcal{E}$. This decomposition reduces the problem of optimizing $J(\pi_\theta;\mathcal{E})$ to three tractable subproblems: (1)~identifying the capabilities that $\pi_\theta$ lacks in $\mathcal{E}$, i.e., those for which $\nu_c(x,\tau)=0$ across many task instances; (2)~synthesizing a capability-targeted environment $\mathcal{E}_s^c$ for each identified capability; and (3)~training $\pi_\theta$ on each $\mathcal{E}_s^c$ to improve performance back in $\mathcal{E}$. In the following subsections, we describe \tool{}, an end-to-end self-improving system that tackles each of these subproblems.

\subsection{Automated Capability-Targeted Synthetic Environment Generation Pipeline}
\label{sec:env_gen}

Given a representative dataset $\mathcal{D}=\{(x_i,\tau_i,r_i,y_i)\}_{i=1}^N$ collected by rolling out $\pi_\theta$ in the target environment $\mathcal{E}$, \tool{} introduces an agentic pipeline that automatically: (i) identifies capabilities that distinguish failed trajectories from successful ones and (ii) constructs a capability-targeted synthetic environment for each.

\noindent\textbf{Contrastive capability identification.}
Given $\mathcal{D}$, our goal is to identify capabilities that $\pi_\theta$ lacks and prioritize those responsible for a substantial fraction of failures, so that they can
be targeted for training. We first split $\mathcal{D}$ into successful and failed subsets according to the
success label $y_i$:
$\mathcal{D}^{+}=\{(x_i,\tau_i,r_i,y_i)\in\mathcal{D}\mid y_i=1\}$ and $\mathcal{D}^{-}=\{(x_i,\tau_i,r_i,y_i)\in\mathcal{D}\mid y_i=0\}$. 

Then, to ensure capability definitions remain consistent across multiple independent analysis runs, we structure the identification process into two phases: discovery and labeling. First, in the \emph{discovery phase}, an LLM-based analysis agent examines the tool calls, tool results, and final responses across the trajectories to induce a canonical dictionary of candidate recurring capabilities $\mathcal{C}$. Each capability $c \in \mathcal{C}$ is assigned a fixed name and natural-language description. Second, in the \emph{labeling phase}, the analysis agent uses this fixed dictionary to systematically evaluate the dataset. For each trajectory $\tau_i$ and capability $c \in \mathcal{C}$, the agent predicts a label $\ell_c(x_i,\tau_i) \in \{\texttt{NA}, \texttt{PRESENT}, \texttt{LACKING}\}$.
Here, \texttt{NA} means the analysis agent predicts $x_i \notin X_c$ ($c$ is not necessary for $x_i$), \texttt{PRESENT} means the analysis agent predicts $x_i \in X_c$ and $\nu_c(x_i,\tau_i)=1$ ($c$ is necessary for $x_i$ and exercised in $\tau_i$), and \texttt{LACKING} means the analysis agent predicts $x_i \in X_c$ and $\nu_c(x_i,\tau_i)=0$ ($c$ is necessary for $x_i$ but not exercised in $\tau_i$). For each capability $c\in\mathcal{C}$, we estimate its error rate on successful
and failed trajectories:

Let $\ell_i^c := \ell_c(x_i,\tau_i)$. Then
\begin{align*}
\widehat{\mathrm{ER}}^{+}(c) &=
\frac{\sum_i \mathbf{1}\!\left[\ell_i^c=\textsc{Lacking},\, y_i=1\right]}
{\sum_i \mathbf{1}\!\left[\ell_i^c\neq\textsc{NA},\, y_i=1\right]}, \\
\widehat{\mathrm{ER}}^{-}(c) &=
\frac{\sum_i \mathbf{1}\!\left[\ell_i^c=\textsc{Lacking},\, y_i=0\right]}
{\sum_i \mathbf{1}\!\left[\ell_i^c\neq\textsc{NA},\, y_i=0\right]}.
\end{align*}

We then define the contrastive gap as
$
\widehat{\Delta}(c)
=
\widehat{\mathrm{ER}}^{-}(c)-\widehat{\mathrm{ER}}^{+}(c).
$ which measures how much more often $c$ is lacking in failed trajectories than in
successful ones. This contrastive criterion is more robust than analyzing failures alone: it filters out capabilities with uniformly low success rates, which often reflect task ambiguity or annotation noise rather than a trainable deficit, and it excludes capabilities that occasionally fail but do not meaningfully distinguish successful from unsuccessful trajectories. 

We also compute the coverage of each capability over failed trajectories $\widehat{\mathrm{Cov}}(c) = 1/|\mathcal{D}^{-}| \sum_{i=1}^{N} \mathbf{1}[\ell_c(x_i,\tau_i)=\textsc{Lacking} \wedge y_i=0]$ and retain $\mathcal{C}^{*} = \{ c \in \mathcal{C} \mid \widehat{\Delta}(c)\ge \delta, \widehat{\mathrm{Cov}}(c)\ge \rho \}$, representing capabilities that are both strongly contrastive and cover a substantial fraction of failures.

To improve robustness, we repeat the analysis except the discovery step multiple times and retain only capabilities selected consistently across runs. In our experiments, one analysis pass takes approximately $5$ minutes, which is marginal compared to training overhead. We set $\rho=0.10$ (a retained capability must account for at least $10\%$ of failed trajectories) and $\delta=0.20$ (its error rate must be at least $20$ percentage points higher on failed than on successful trajectories). We analyze and provide examples of the identified capabilities in Section~\ref{sec:env_analysis} and Appendix~\ref{app:cap_det}.

\noindent\textbf{Environment synthesis.}
For each retained capability $c \in \mathcal{C}^{*}$, an LLM-based generation agent synthesizes a complete training environment $\mathcal{E}_s^c$. The agent receives the description of $c$ together with the failure patterns identified during contrastive analysis, and produces a seeded task generator $G_c$, transition logic, and evaluation criteria that together define $\mathcal{E}_s^c$ while preserving the target environment's tool schemas, state representation, and policy constraints. Given a random seed $z$, the generator deterministically constructs a task instance $x = G_c(z)$, including synthetic user profiles, database records, and task-specific parameters, so that each seed yields a distinct scenario, preventing memorization and enabling diversity.

By construction, task success in $\mathcal{E}_s^c$ depends primarily on exercising $c$, which is verified automatically from tool arguments, state changes, or final outputs. For example, if $c$ is \emph{structured data reasoning}, the environment generates scenarios requiring the agent to search, filter, and cross-reference specific records within complex JSON databases (such as locating a specific flight routing or matching a retail item variant). For each seeded instance, the environment also generates a corresponding ground-truth solution. For that instance, the agent interacts in a multi-turn setting, using tools to query records, modify the database, and converse with a simulated user. The reward is computed by comparing the final database state against the ground truth using hash-based consistency checks and by verifying that the agent communicates the correct final result. Because each environment isolates a single capability, the reward $R_{\mathcal{E}_s^c}$ is denser and more attributable than that of the original environment, where success may depend on many capabilities at once. Examples of synthesized environments and the generation prompt are in Appendix~\ref{app:env_prompt} and Appendix~\ref{app:synth_env_examples}.

\subsection{Acquiring Capabilities via Reinforcement Learning}
\label{sec:training}
Given the family of synthetic environments $\{\mathcal{E}_s^c\}_{c \in \mathcal{C}^{*}}$ produced by the generation pipeline (\S\ref{sec:env_gen}), we train a separate low-rank adapter $\Delta_c$ ~\citep{hu2022lora} for each capability $c \in \mathcal{C}^{*}$ while keeping the base policy $\pi_\theta$ frozen.

We optimize each adapter with GRPO, a value-free on-policy algorithm. At each iteration, $\pi_{\theta+\Delta_c}$ generates $G$ groups of rollouts in $\mathcal{E}_s^c$. Within each group $g$, $K$ trajectories $\{\tau_{g,1}, \dots, \tau_{g,K}\}$ are sampled from the same seed $z_g$, so all rollouts share an identical initial state and differ only through stochastic decoding. Let $r_{g,k} = R_c(x_g, \tau_{g,k})$ denote the reward function, where $x_g = G_c(z_g)$. GRPO normalizes rewards within each group to obtain a trajectory-level advantage: $\hat{A}_{g,k} = \frac{r_{g,k} - \bar{r}_g}{\sigma_g + \epsilon}$, 
where $\bar{r}_g$ and $\sigma_g$ are the within-group mean and standard deviation and $\epsilon$ is a small constant for numerical stability. This normalization makes the training signal invariant to reward scale across environments. Because reward is assigned at the trajectory level, all tokens in a rollout share the same advantage; groups in which all rollouts receive identical rewards are discarded, as they carry no learning signal. The adapter is then updated via the standard clipped surrogate objective~\citep{schulman2017proximalpolicyoptimizationalgorithms}; full details are provided in Appendix~\ref{app:grpo}.

\subsection{Composing Acquired Capabilities}
\label{sec:composition}

The target environment $\mathcal{E}$ may require different capabilities across task instances. We consequently train a Mixture-of-Experts (MoE) model over the capability adapters.

We construct the MoE by composing the capability-specific adapters
$\{\Delta_c\}_{c \in \mathcal{C}^*}$ at each adapted transformer block using token-level routing ~\cite{dou2024loramoerevolutionizingmixture, wu2024mixtureofloraexperts, shazeer2017outrageouslylargeneuralnetworks}.
At each adapted block $\ell$, we stack the LoRA factors
$\{(A_c, B_c)\}_{c \in \mathcal{C}^*}$ into a shared adapter bank and introduce a lightweight linear gate $g_\ell$.
For capability $c$, let $\Delta W_c = B_c A_c$ denote its low-rank weight update.
Given hidden state $h$ at block $\ell$, the gate produces capability weights
$p_{\ell,c}(h) = \mathrm{softmax}(g_\ell(h))_c$.
For an adapted projection with frozen base weight $W$, the token-dependent update is
\[
\Delta W_{\ell}(h) = \sum_{c \in \mathcal{C}^*} p_{\ell,c}(h)\,\Delta W_c,
\]
and the projection output is $(W + \Delta W_{\ell}(h))h$.
The base policy $\pi_\theta$ and all adapters are frozen; only the gates $\{g_\ell\}$ are trained.
We train the gates with the objective
\[
\mathcal{L}
=
\mathcal{L}_{\mathrm{CE}}
+
\lambda_{\mathrm{LB}}\,\mathcal{L}_{\mathrm{LB}},
\]
where $\mathcal{L}_{\mathrm{CE}}$ is the token-level cross-entropy on the agent's response tokens and $\mathcal{L}_{\mathrm{LB}} = |\mathcal{C}^*|\sum_{c \in \mathcal{C}^*} f_c\, P_c$ is a load-balancing penalty that discourages expert collapse. Here, $f_c$ is the fraction of batch tokens for which $c$ has the largest gate probability, $P_c$ is the average gate probability assigned to $c$, and $\lambda_{\mathrm{LB}}$ controls the strength of the regularizer. At inference, each token uses the single capability with the largest gate probability. We empirically compare this MoE composition against alternative consolidation strategies (adapter merging, multi-capability GRPO, SFT, and on-policy distillation) in Appendix~\ref{app:capability_consolidation}.

\section{Experiments}

\subsection{Experimental Setup}

\noindent\textbf{Benchmarks \& Evaluation.}
We evaluate on \tbench{}~\citep{barres2025tau2benchevaluatingconversationalagents} (using the 50 \textsc{Airline} and 114 \textsc{Retail} domains) to test policy-sensitive customer-service workflows, and \swe{}~\citep{jimenez2024swebenchlanguagemodelsresolve} (500 instances) to test software-engineering tasks that require resolving real GitHub issues. On \tbench{}, we report \emph{pass rate}, the fraction of tasks satisfying the benchmark's binary success criterion (correct task execution and correct user communication), broken out by the \textsc{Airline} and \textsc{Retail} domains and aggregated over the union of both. On \swe{}, we report \emph{Pass@1}, the fraction of instances whose generated patch resolves the issue and passes the held-out tests; all methods, including \tool{}'s capability discovery and synthetic environment generation, train on the disjoint SWE-smith~\citep{yang2025swesmithscalingdatasoftware} split. Formal metric definitions are in Appendix~\ref{app:metrics}. We additionally evaluate on \tsb{}~\citep{lu2025toolsandboxstatefulconversationalinteractive}, a stateful tool-use benchmark, in Appendix~\ref{app:toolsandbox}. Unless specified otherwise, we evaluate all models with greedy decoding (temperature $0.0$) for deterministic head-to-head comparison, a maximum context length of 32{,}000 tokens, and a maximum of 50 interaction steps per episode.

\noindent\textbf{Training Hyperparameters.}
We evaluate with two agent backbones, Qwen3-30B-A3B-Instruct-2507~\citep{qwen3technicalreport} and Qwen3.6-27B~\citep{qwen3.6-27b} (Table~\ref{tab:main_tb_results}), optimizing with LoRA and GRPO. On \tbench, the 4 capabilities trained are structured data reasoning, multi-step task completion, precondition verification, and tool calling precision. On \swe, we train 3 capability-targeted synthetic environments. We optimize with AdamW ($\text{lr}=10^{-5}$) for up to 40 iterations per capability, using a sampling temperature of $1.0$ for rollouts. Training utilizes gradient checkpointing and distributed data parallelism across 4--8 A100-80GB GPUs. MoE training uses a load balancing loss coefficient $\lambda_{\mathrm{LB}} = 0.05$. As detailed in Appendix~\ref{app:param_counts}, each LoRA adapter introduces $\sim\!1.6$\,B trainable parameters ($5.3\%$ of the backbone), and the trained MoE gate adds $491{,}760$ parameters in total.

\noindent\textbf{Baselines.}
We compare against direct GRPO training on the target environment, \emph{Agent World Model (AWM)}, which performs RL in benchmark-independent synthetic environments for broad tool-use training, and \emph{Agent Data Protocol (ADP)}, which performs supervised fine-tuning on diverse public agent trajectories in a unified representation. On \swe{}, we additionally compare against \emph{SWE-RL}~\citep{wei2025swerl}, a reinforcement learning method specialized for software-engineering agents.

To test whether dedicated training on the identified capabilities is necessary, we compare against the inference-time evolutionary prompt optimization method GEPA. We provide the capabilities identified by \tool{} to GEPA and allow it to refine their natural-language descriptions, which are then added to the agent's system prompt. GEPA optimizes these descriptions for the agent's pass rate in the target environment. All training baselines and GEPA are run under matched training budgets to ensure fair comparison with \tool{}.

\begin{table}[t]
\small
\centering
\caption{Pass rate on \tbench{} with two agent backbones. \tool{} outperforms all baselines under both Qwen3-30B-A3B and Qwen3.6-27B.}
\label{tab:main_tb_results}
\setlength{\tabcolsep}{4pt}
\begin{tabular}{@{}llccc@{}}
\toprule
\textbf{Model} & \textbf{Method} & \textbf{Airline (\%)} & \textbf{Retail (\%)} & \textbf{Overall (\%)} \\
\midrule
Qwen3-30B-A3B & Base & 24.0 & 36.8 & 32.9 \\
              & GRPO on Target & 32.0 & 40.4 & 37.8 \\
              & ADP & 28.0 & 34.2 & 32.3 \\
              & AWM & 32.0 & 41.2 & 38.4 \\
              & GEPA & 38.0 & 40.4 & 39.6 \\
              & Single Capability GRPO (Ours) & 34.0 & 43.0 & 40.3 \\
              & \textbf{TRACE} (Ours) & \textbf{44.0} & \textbf{50.0} & \textbf{48.2} \\
\midrule
Qwen3.6-27B   & Base & 60.0 & 45.6 & 50.0 \\
              & GRPO on Target & 66.0 & 48.2 & 53.7 \\
              & ADP & 56.0 & 49.1 & 51.2 \\
              & GEPA & 66.0 & 47.4 & 53.0 \\
              & Single Capability GRPO (Ours) & 66.0 & 50.8 & 55.4 \\
              & \textbf{TRACE} (Ours) & \textbf{74.0} & \textbf{52.6} & \textbf{59.1} \\
\bottomrule
\end{tabular}
\end{table}

\begin{table}[t]
\small
\centering
\caption{Pass@1 on \swe{} (500 instances) with two agent backbones.}
\label{tab:swe_results}
\setlength{\tabcolsep}{4pt}
\begin{tabular}{@{}llc@{}}
\toprule
\textbf{Model} & \textbf{Method} & \textbf{Pass@1 (\%)} \\
\midrule
Qwen3-30B-A3B & Base & 26.0 \\
              & ADP & 30.2 \\
              & GEPA & 31.0 \\
              & SWE-RL & 32.6 \\
              & Single Capability GRPO (Ours) & 36.6 \\
              & \textbf{TRACE} (Ours) & \textbf{41.0} \\
\midrule
Qwen3.6-27B   & Base & 68.0 \\
              & ADP & 68.4 \\
              & GEPA & 69.6 \\
              & SWE-RL & 70.4 \\
              & Single Capability GRPO (Ours) & 71.0 \\
              & \textbf{TRACE} (Ours) & \textbf{73.2} \\
\bottomrule
\end{tabular}
\end{table}

\begin{figure}[!t]
    \centering
    \begin{subfigure}[t]{0.48\textwidth}
        \centering
        \includegraphics[width=\linewidth]{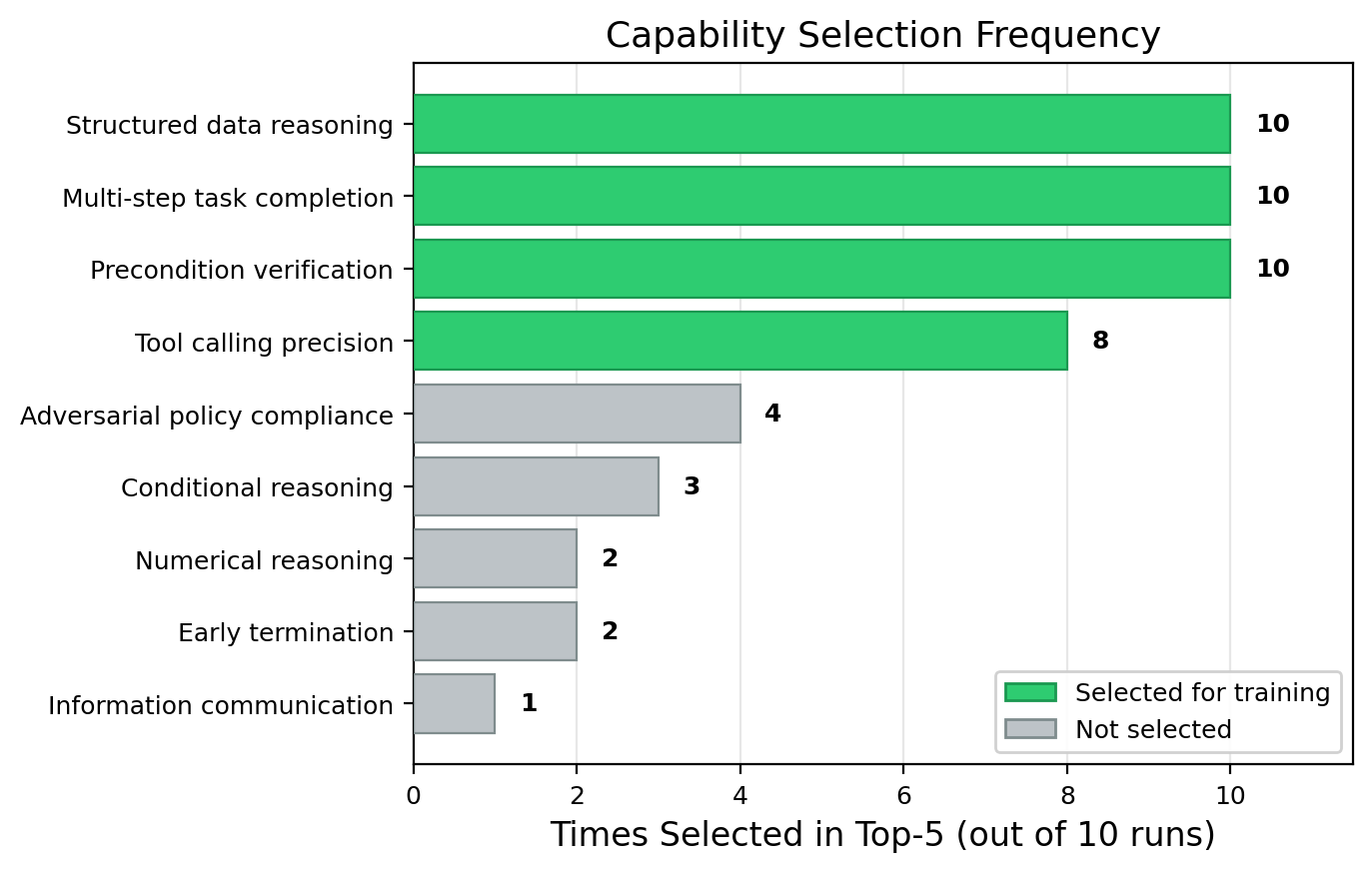}
        \caption{Capability selection frequency across 10 independent runs. The analysis agent consistently recovers the same top four deficits (green) for targeted training.}
        \label{fig:capability_selection}
    \end{subfigure}
    \hfill
    \begin{subfigure}[t]{0.48\textwidth}
        \centering
        \includegraphics[width=\linewidth]{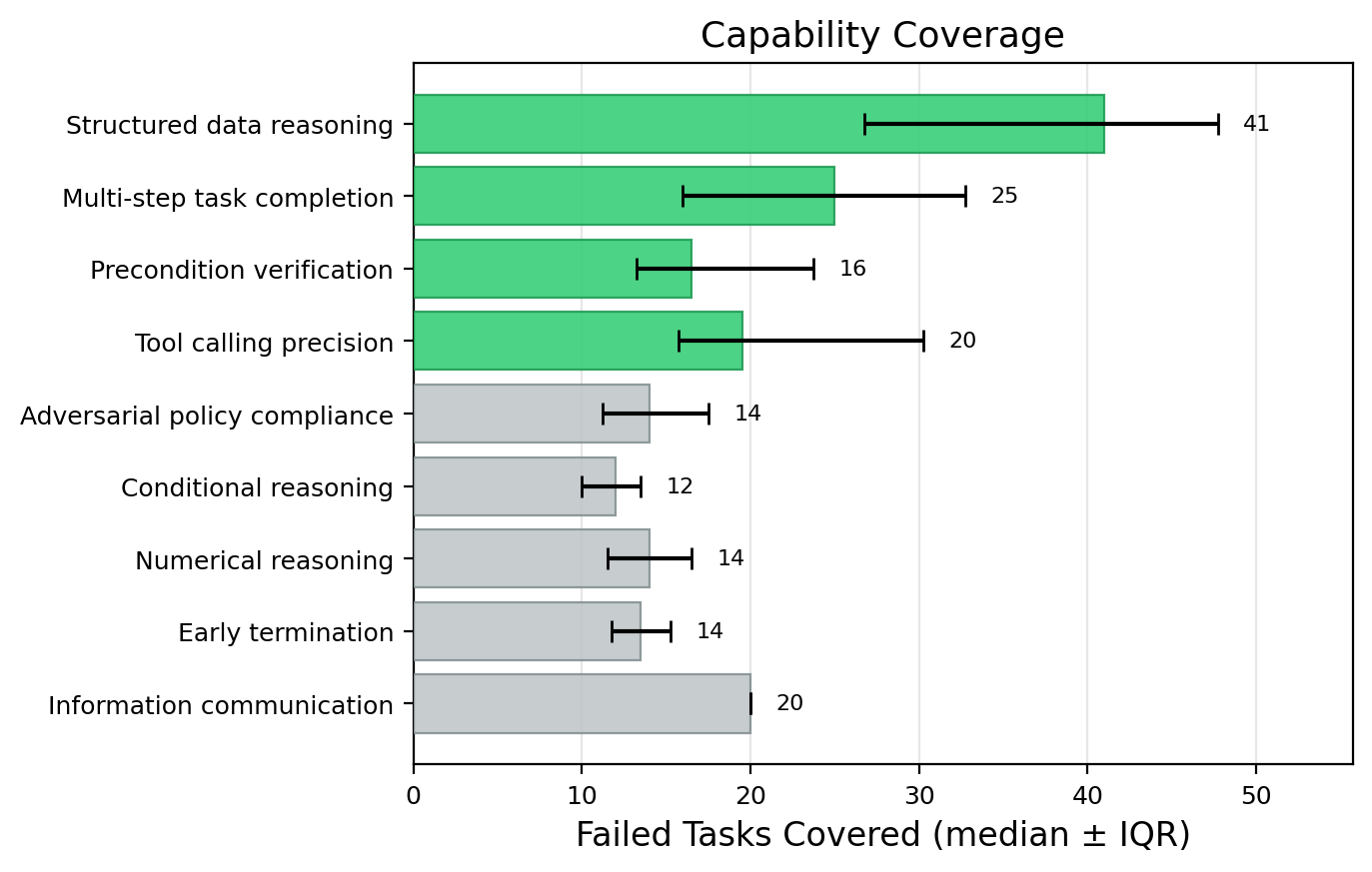}
        \caption{Median task failure coverage across runs (error bars denote IQR). Coverage is not mutually exclusive, as a single failed task may lack multiple capabilities.}
        \label{fig:capability_coverage}
    \end{subfigure}
    \caption{Stability and coverage of identified capabilities. Repeated contrastive analysis consistently identifies a small, stable set of capability deficits (a) that account for a heavily concentrated proportion of benchmark failures (b). This skewed distribution demonstrates that a few high-impact capabilities drive most errors, validating our use of targeted synthetic environments.}
    \label{fig:capability_analysis}
\end{figure}

\subsection{Effectiveness of Method}

Tables~\ref{tab:main_tb_results} and~\ref{tab:swe_results} summarize the main experimental results. \tool{} consistently outperforms baselines across both evaluated benchmarks, notably improving upon the base model by +15.3 points on \tbench{} and +15 points Pass@1 on \swe{}, while beating the strongest external baselines by +8.6 points and +8.4 points, respectively. The same pattern holds with a stronger Qwen3.6-27B backbone on \tbench{} (Table~\ref{tab:main_tb_results}): \tool{} reaches a 59.1\% overall pass rate, +9.1 points over the base model and +5.4 points over the strongest baseline. On \swe{} Verified, the Qwen3.6-27B variant of \tool{} reaches 73.2\% Pass@1 (Table~\ref{tab:swe_results}), outperforming much larger frontier models, including GPT-5.2-Codex, GLM 5, and Claude 4.5 Sonnet on the public SWE-bench leaderboard.

\noindent\textbf{Targeted vs. General-Purpose Training Environments.}
A single adapter trained on one synthesized capability environment achieves 40.3\% on \tbench{} and 36.6\% Pass@1 on \swe{}. This surpasses large-scale general-purpose methods such as AWM (38.4\% on \tbench{}) and ADP (32.3\% on \tbench{}, 30.2\% on \swe{}). Notably, ADP's performance in the \tbench{} Retail domain is lower than the base model (34.2\% vs.\ 36.8\%). These results suggest that diagnosing specific missing capabilities and training on targeted synthetic environments yields higher performance than general-purpose training data.

\noindent\textbf{Capability Training vs. Prompt Optimization.}
We evaluate the need for dedicated training versus injecting capability descriptions into the prompt optimized with GEPA. As shown in Tables~\ref{tab:main_tb_results} and~\ref{tab:swe_results}, GEPA improves over the base model on both \tbench{} (overall pass rate) and \swe{} (Pass@1). However, it underperforms compared to training a single capability and \tool{} with all capabilities on both benchmarks. This demonstrates that while prompting capability instructions is helpful, explicitly training the model to exercise those capabilities is necessary for stronger performance.

\subsection{Analysis of Synthesized Environments}
\label{sec:env_analysis}

We analyze the capabilities identified by repeated contrastive analysis across 10 independent runs (Figure~\ref{fig:capability_analysis}a). The pipeline consistently converges on a small set of lacking capabilities: structured data reasoning, multi-step task completion, and precondition verification are recovered in all 10 runs, while tool-calling precision appears in 8 of 10.

Beyond selection frequency, failure coverage is heavily concentrated within these specific capabilities (Figure~\ref{fig:capability_analysis}b). Structured data reasoning accounts for the largest share of failed benchmark tasks, followed closely by multi-step task completion. This skewed distribution empirically validates our approach of explicitly isolating and targeting the lacking capabilities responsible for a significant proportion of failures, which maximizes data efficiency. More detailed analysis is provided in Appendix~\ref{sec:appendix_capability}.

\begin{figure}[!h]
    \centering
    \begin{subfigure}[t]{0.48\textwidth}
        \centering
        \includegraphics[width=\linewidth]{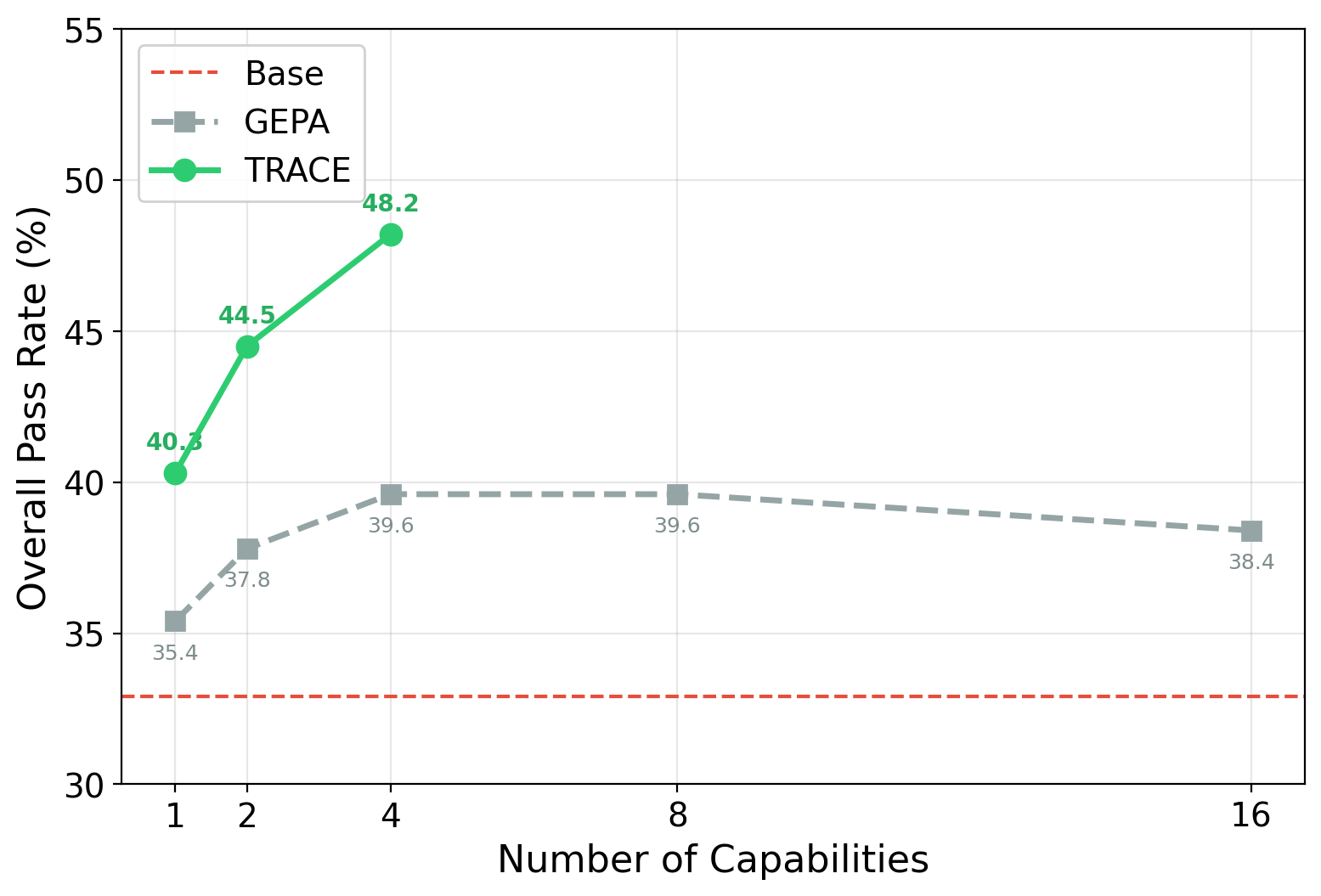}
        \caption{Scaling the number of capabilities (\tool{} vs.\ GEPA).}
        \label{fig:cap_scaling}
    \end{subfigure}
    \hfill
    \begin{subfigure}[t]{0.48\textwidth}
        \centering
        \includegraphics[width=\linewidth]{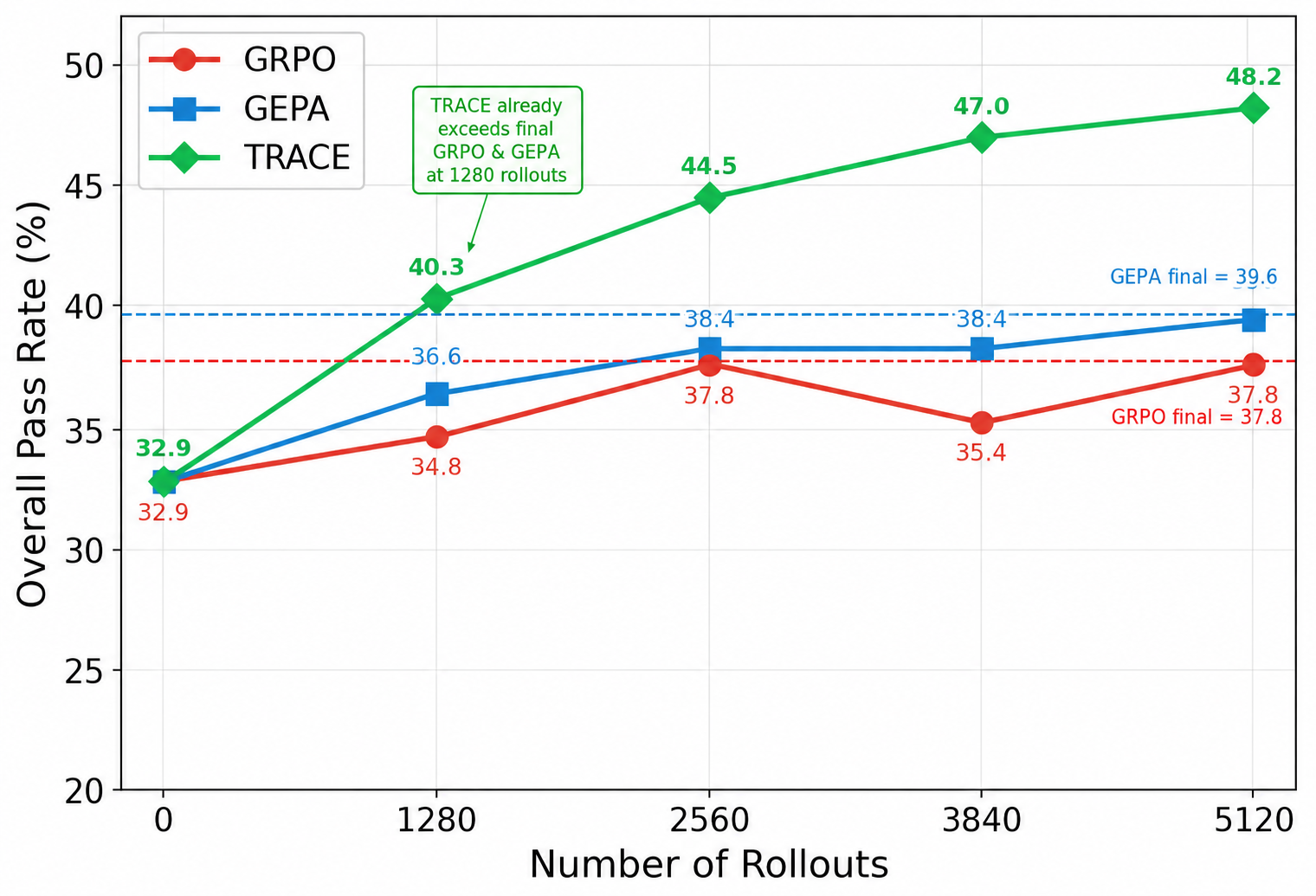}
        \caption{Scaling the number of rollouts.}
        \label{fig:scaling_rollouts}
    \end{subfigure}
    \caption{Method scaling on \tbench{}. (a) \tool{} overall pass rate as the number of trained capabilities increases, compared to GEPA prompt optimization. (b) Pass rate as the number of rollouts increases for \tool{}, GEPA, and GRPO on the target environment.}
    \label{fig:method_scaling}
\end{figure}

\subsection{Method Scaling}
\label{sec:scaling}

Figure~\ref{fig:cap_scaling} shows how \tool{}'s overall pass rate scales with the number of trained capabilities. In contrast, GEPA, which optimizes capability descriptions in the prompt, plateaus after 4 capabilities. Figure~\ref{fig:scaling_rollouts} illustrates how performance on \tbench{} scales with the number of rollouts for \tool{}, GEPA, and running GRPO on the target environment. \tool{} demonstrates consistent, monotonic improvement, scaling from a base pass rate of 32.9\% up to 48.2\% at 5{,}120 rollouts. In contrast, GEPA plateaus early, reaching only 39.6\%, while GRPO exhibits instability, dropping to 35.4\% at 3{,}840 rollouts, and ultimately stalls at 37.8\%. A consistent trend appears on \tsb{} (Appendix~\ref{app:toolsandbox}).

\section{Conclusion}
In this work, we propose \tool{}, an end-to-end system for environment-specific agent self-improvement that automatically identifies lacking capabilities from the agent's trajectories, synthesizes targeted training environments for each capability, trains capability-specific LoRA adapters via RL, and then trains a Mixture-of-Experts (MoE) model over the capability adapters. \tool{} can be effectively applied across different environments, improving over the base agent by \textbf{+15.3 points} on \tbench{} (customer service) and \textbf{+15 points} Pass@1 on \swe{} (software engineering), outperforming the strongest external baseline by \textbf{+8.6 points} and \textbf{+9 points}, respectively. Given the same number of rollouts, \tool{} scales more efficiently than baselines, outperforming GRPO and GEPA by \textbf{+10.4} and \textbf{+8.6} points on \tbench{}, respectively.

\section*{Acknowledgements}
We thank the Scaling Intelligence Lab and others for their constructive feedback during the composition of the paper. In particular, we would like to thank Debangshu Banerjee, Tanvir Bhathal, Alex Bloom, Andy Dimnaku, Simon Guo, Sid Jha, Hermann Kumbong, Jacky Kwok, Andrew Shi, and Shayan Talaei. We also thank Prime Intellect, Lambda Labs, Google, Modal, and IBM for providing compute resources that enabled our experiments.
\newpage
\bibliography{colm2026_conference}
\bibliographystyle{colm2026_conference}

\newpage
\appendix

\section{GRPO Details}
\label{app:grpo}
Given the family of synthetic environments $\{\mathcal{E}_s^c\}_{c \in \mathcal{C}^{*}}$ produced by the environment generation pipeline (\S\ref{sec:env_gen}), we train a separate low-rank adapter~\citep{hu2022lora} $\Delta_c$ for each capability $c \in \mathcal{C}^{*}$ while keeping the base model $\pi_\theta$ frozen.

\noindent\textbf{GRPO training.} We train with Group Relative Policy Optimization, an on-policy reinforcement learning algorithm that avoids learning an explicit value function. For each capability $c$, the adapted policy $\pi_{\theta+\Delta_c}$ interacts with the corresponding synthetic environment $\mathcal{E}_s^c$. At each training iteration, the policy generates $G$ groups of rollouts. Within each group $g$, $K$ trajectories $\{\tau_{g,1}, \dots, \tau_{g,K}\}$ are sampled from the same environment seed $z_g$, so that all $K$ rollouts begin from an identical initial state and differ only through stochastic decoding. Let $r_{g,k} = R_{\mathcal{E}_s^c}(x_g, \tau_{g,k})$ denote the reward assigned by the programmatic verifier of $\mathcal{E}_s^c$, where $x_g = G_{\mathcal{E}_s^c}(z_g)$. GRPO computes a group-relative normalized advantage for each trajectory,
\begin{equation}
\hat{A}_{g,k} = \frac{r_{g,k} - \bar{r}_g}{\sigma_g + \epsilon},
\qquad
\bar{r}_g = \frac{1}{K}\sum_{j=1}^{K} r_{g,j},
\qquad
\sigma_g = \sqrt{\frac{1}{K}\sum_{j=1}^{K}(r_{g,j}-\bar{r}_g)^2},
\end{equation}
where $\bar{r}_g$ and $\sigma_g$ are the within-group mean and standard deviation. This normalization makes the training signal invariant to reward scale across capabilities and synthetic environments. Because reward is assigned at the trajectory level, all action tokens in a rollout receive the same advantage. Groups in which all $K$ rollouts receive identical rewards are discarded, since they provide no learning signal.

The adapter parameters are updated using the clipped GRPO surrogate objective
\begin{equation}
\mathcal{L}_{\mathrm{GRPO}}
=
-\frac{1}{|\mathcal{B}|}
\sum_{(g,k)\in\mathcal{B}}
\frac{1}{T_{g,k}}
\sum_{t=1}^{T_{g,k}}
\min\!\Bigl(
\rho_t \hat{A}_{g,k},
\mathrm{clip}(\rho_t, 1-\epsilon, 1+\epsilon)\hat{A}_{g,k}
\Bigr),
\label{eq:grpo}
\end{equation}
where
\[
\rho_t
=
\frac{\pi_{\theta+\Delta_c}(a_t \mid x, a_{<t})}
{\pi_{\theta+\Delta_c}^{\mathrm{old}}(a_t \mid x, a_{<t})}
\]
is the per-token importance ratio between the current and rollout policies, $a_t$ denotes the $t$-th token in the trajectory, $T_{g,k}$ is the number of tokens in trajectory $\tau_{g,k}$, $\mathcal{B}$ denotes the minibatch, and $\epsilon$ is the clipping threshold. As in PPO-style methods, clipping prevents excessively large policy updates when the estimated advantage is high.

\noindent\textbf{On-policy rollout collection.} At each iteration, we synchronize the current adapter $\Delta_c$ to a separate inference worker and collect rollouts on-policy from $\pi_{\theta+\Delta_c}$. The policy interacts with the task generator $G_{\mathcal{E}_s^c}$, receives observations, invokes tools, and continues until the episode terminates, after which the reward function $R_{\mathcal{E}_s^c}$ assigns the terminal reward. Because rollouts are collected on-policy, the importance ratio $\rho_t$ remains close to $1$ at the beginning of each update, which improves optimization stability.
\newpage

\section{ToolSandbox Evaluation}
\label{app:toolsandbox}

We additionally evaluate \tool{} on \tsb{}~\citep{lu2025toolsandboxstatefulconversationalinteractive} (129 base scenarios), a stateful, conversational tool-use benchmark. We report \emph{mean similarity}, the average $[0,1]$ trajectory score from the benchmark's milestone-based evaluator, and \emph{perfect rate}, the fraction of scenarios scoring exactly $1.0$ (formal definitions in Appendix~\ref{app:metrics}). For \tool{}, the two capabilities trained are permission error recovery and datetime reasoning (Appendix~\ref{app:cap_det}). As in the main paper, all baselines are run under matched training budgets and evaluated with greedy decoding.


\begin{table}[t]
\small
  \centering
  \caption{Perfect Score (similarity $= 1.0$) and Mean Similarity on \tsb}
  \label{tab:toolsandbox}
  \begin{tabular}{lcc}
  \toprule
  \textbf{Model} & \textbf{Perfect} & \textbf{Mean Sim.} \\
  \midrule
  Base & 19/129 & 0.411 \\
  ADP &     19/129  &   0.422 \\
  GRPO on Target & 22/129 & 0.519 \\
  AWM & 20/129 & 0.504 \\
  GEPA & 22/129 & 0.520 \\
  Single Capability GRPO (Ours) &  22/129 & 0.514 \\
  \textbf{TRACE} (Ours) & \textbf{26 / 129} & \textbf{0.560} \\
  \bottomrule
  \end{tabular}
  \end{table}

Table~\ref{tab:toolsandbox} summarizes the results. \tool{} achieves a mean similarity of $0.560$ with $26$ perfect scores, improving over the base model by $+0.149$ mean similarity and $7$ additional perfect scores, and over the strongest external baseline by $+0.040$ mean similarity and $4$ additional perfect scores. A single adapter trained on one synthesized capability environment already reaches $0.514$, surpassing the general-purpose AWM ($0.504$) and ADP ($0.422$) baselines, while prompt optimization with GEPA reaches $0.520$. These trends mirror those on \tbench{} and \swe{}: diagnosing and training the specific missing capabilities outperforms both general-purpose training data and prompt optimization.

\begin{figure}[t]
    \centering
    \includegraphics[width=0.55\textwidth]{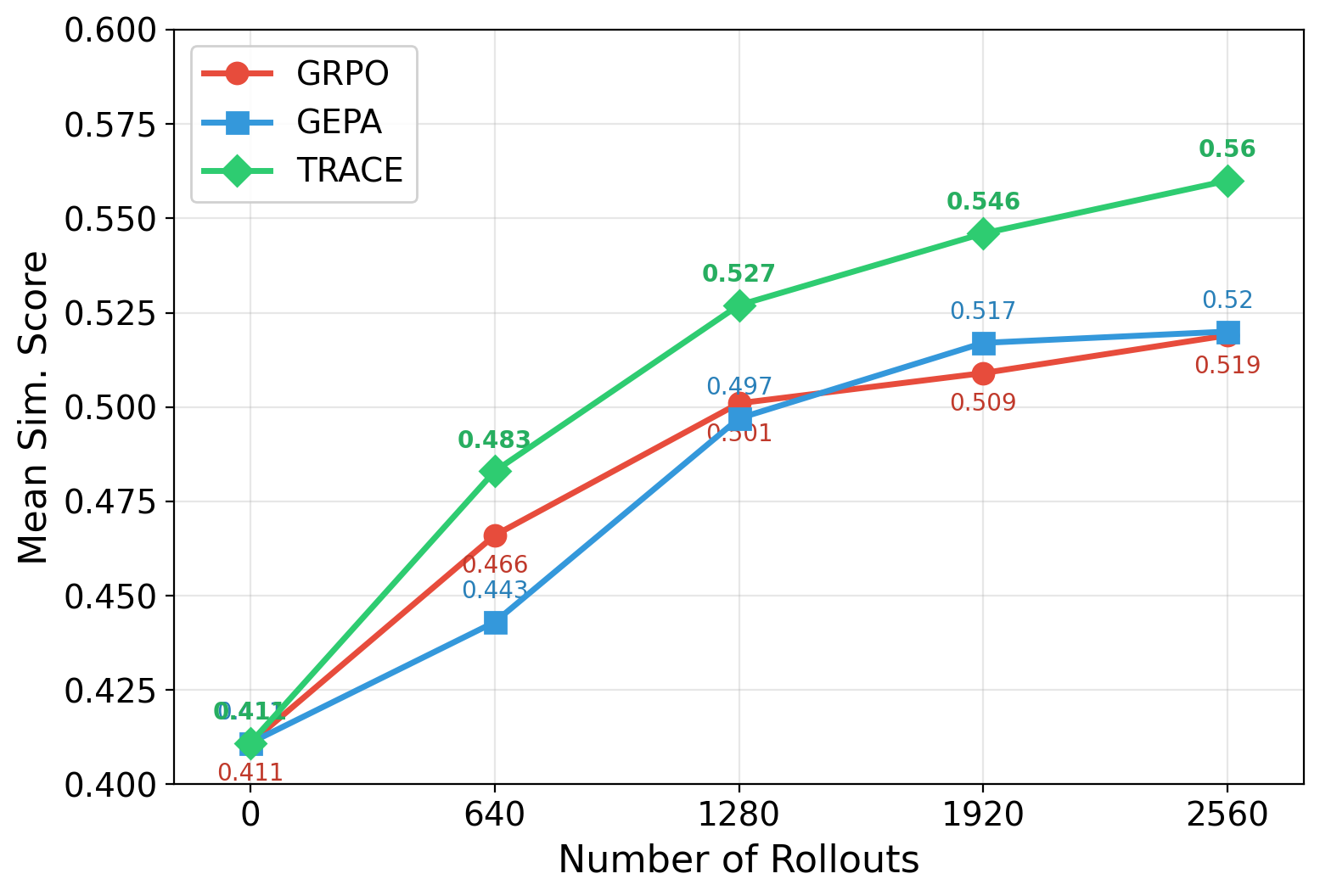}
    \caption{Mean similarity on \tsb{} as the number of rollouts is scaled. As on \tbench{}, \tool{} improves steadily to a $0.560$ mean similarity score, outperforming both GEPA ($0.520$) and GRPO ($0.519$).}
    \label{fig:scale_rollouts_tsb}
\end{figure}

\section{Capability Consolidation}
\label{app:capability_consolidation}

\begin{table}[t]
\small
\centering
\caption{Single Model Consolidation Approaches on \tbench}
\label{tab:multi_tb_ablation}
\setlength{\tabcolsep}{2pt}
\begin{tabular}{@{}lccc@{}}
\toprule
\textbf{Approach} & \textbf{Airline (\%)} & \textbf{Retail (\%)} & \textbf{Overall (\%)} \\
\midrule
Base & 24.0 & 36.8 & 32.9 \\
Single Capability GRPO & 34.0 & 43.0 & 40.3 \\
CORE-TSV & 36.0 & 41.2 & 39.6 \\
On Policy Distillation & 28.0 & 42.1 & 37.8 \\
SFT Synthetic & 36.0 & 38.6 & 37.8 \\
Multi Capability GRPO & 38.0 & 42.1 & 40.9 \\
\textbf{TRACE} & \textbf{44.0} & \textbf{50.0} & \textbf{48.2} \\
\bottomrule
\end{tabular}
\end{table}

In Table~\ref{tab:multi_tb_ablation}, we compare \tool{} MoE against alternatives to consolidate capabilities into one model. \emph{CORE-TSV merge}~\citep{panariello2025corespace,ilharco2023taskarithmetic,yadav2023tiesmerging} combines independently trained, capability-specific adapters in a shared low-rank space using Core Space merging, task-vector composition, and TIES-style sign resolution. \emph{Multi-Capability GRPO} trains a single LoRA adapter on a uniform mixture of all synthesized capability environments. \emph{SFT Synthetic} trains a single LoRA adapter on successful trajectories collected from all the synthesized capability environments. \emph{On-Policy Distillation} trains a LoRA teacher for each synthetic capability environment, then trains one student LoRA on a uniform mix by having the student generate on-policy rollouts and match the relevant teacher’s token-level distributions~\citep{lu2025onpolicydistillation}. All consolidation approaches are trained under matched budgets. Only Multi-Capability GRPO (40.9\%) marginally outperforms the best single adapter (40.3\%), whereas MoE significantly outperforms it.




\section{Parameter Counts}
\label{app:param_counts}

\begin{table}[t]
\centering
\small
\caption{Parameter counts for \tool{} Model.}
\label{tab:moe-gate}
\begin{tabular}{lr}
\toprule
Component & Number of Parameters \\
\midrule
Qwen3-30B-A3B backbone (frozen) & $\sim\!30.5$\,B \\
Capability LoRA adapter (each) & $\sim\!1.6$\,B \\
MoE gate (Linear $d{\to}|\mathcal{C}^*|$, one per block) & 491{,}760 \\
\bottomrule
\end{tabular}
\end{table}

Table~\ref{tab:moe-gate} breaks down the parameters of each \tool{} component. The Qwen3-30B-A3B backbone ($\sim\!30.5$\,B parameters) remains frozen throughout the entire pipeline. Each capability-specific LoRA adapter introduces $\sim\!1.6$\,B trainable parameters (corresponding to $5.3\%$ of the backbone) and is optimized independently on its respective synthetic environment during the RL stage (\S\ref{sec:training}). At the composition stage (\S\ref{sec:composition}), the backbone and all $|\mathcal{C}^*|$ capability adapters are frozen, and the only trainable parameters are the MoE gates: one linear map from the block's hidden dimension $d$ to the $|\mathcal{C}^*|$ capability logits per adapted transformer block, totaling $491{,}760$ parameters across all blocks.

Although \tool{} stores all $|\mathcal{C}^*|$ adapters, inference is sparse: at each token, the MoE gate selects the top-$k$ capabilities, so only $k$ of the adapters are active. In our experiments we find $k{=}1$ works well, holding the per-token active parameter count to the backbone's active footprint plus a single adapter.

\section{Environment Generation Prompt}
\label{app:env_prompt}
\lstset{
    basicstyle=\ttfamily\footnotesize,
    breaklines=true,
    keywordstyle=\color{blue},
    commentstyle=\color{green!60!black},
    stringstyle=\color{orange},
    showstringspaces=false
}

\newtcolorbox{promptbox}[1][]{
    enhanced,
    breakable,
    colback=gray!5!white,
    colframe=gray!75!black,
    title=System Prompt: RL Environment Designer,
    fonttitle=\bfseries\sffamily,
    attach boxed title to top left={xshift=5mm, yshift=-2mm},
    boxed title style={colback=gray!75!black, sharp corners=all, rounded corners=northeast, rounded corners=northwest},
    top=15pt,
    bottom=10pt,
    left=10pt,
    right=10pt,
    #1
}

\begin{promptbox}

\noindent You are an RL environment designer. Your job is to create a GRPO-compatible training environment that teaches a language model a specific skill through procedurally generated scenarios with shaped rewards.

\vspace{1em}
\noindent \textbf{\large Target Skill}
\begin{itemize}
    \item \{SKILL\_DESCRIPTION\}
    \item What the model currently does wrong: \{FAILURE\_PATTERN\}
    \item What correct behavior looks like: \{CORRECT\_BEHAVIOR\}
    \item Example trajectories showing the failure: \newline \{TRAJECTORY\_EXAMPLES\_OR\_PATH\}
\end{itemize}

\vspace{1em}
\noindent \textbf{\large Architecture Requirements}

\noindent The environment MUST implement the \texttt{GameEnv} protocol from \texttt{game\_registry.py}:
\begin{lstlisting}[language=Python]
class GameEnv(Protocol):
    done: bool                           # Episode terminated?
    current_player: int                  # Which agent acts next (0 = agent)
    rewards: Dict[int, float]            # Terminal rewards per player
    invalid_player: Optional[int]        # Invalid move tracking

    def reset(self, seed: int) -> None: ...
    def observe(self, player_id: int) -> str: ...
    def legal_actions(self) -> List[str]: ...
    def step(self, action: Optional[str]) -> None: ...
\end{lstlisting}

\noindent And register itself via \texttt{GameSpec} in \texttt{game\_registry.py}:
\begin{lstlisting}[language=Python]
register_game(GameSpec(
    name="{GAME_NAME}",
    make_env=lambda **kwargs: YourGameEnv(**kwargs),
    extract_action=extract_action_fn,
    system_prompt="...",
    max_gen_tokens=2048,
))
\end{lstlisting}

\vspace{1em}
\noindent \textbf{\large Critical Design Constraints}

\vspace{0.5em}
\noindent \textbf{1. Reward Design for GRPO}\\
GRPO learns from WITHIN-GROUP reward variance. If all rollouts in a group get the same reward, the gradient is zero and the iteration is wasted. Your reward function MUST create variance:
\begin{itemize}
    \item \textbf{DO}: Use continuous/multi-level rewards (0.0, 0.2, 0.5, 0.8, 1.0) so that partial success is distinguishable from total failure
    \item \textbf{DO}: Design scenarios where the base model succeeds $\sim$30-60\% of the time. Too easy ($>$80\%) = no negative signal. Too hard ($<$15\%) = no positive signal.
    \item \textbf{DO}: Use multiplicative reward components for behaviors that must co-occur (e.g., \texttt{tool\_correct * communicated\_result}), with an additive floor to preserve gradient signal
    \item \textbf{DON'T}: Use pure binary rewards (0 or 1) unless the base model success rate is $\sim$40-60\%
\end{itemize}

\noindent Recommended reward structure:
\begin{lstlisting}[language=Python]
# Multiplicative component: all must succeed for full reward
mult = component_a * component_b * component_c

# Additive floor: preserves gradient even when one component is zero
add = w_a * component_a + w_b * component_b + w_c * component_c

# Combined: multiplicative dominates but additive prevents zero-gradient
reward = alpha * mult + (1 - alpha) * add
\end{lstlisting}

\vspace{0.5em}
\noindent \textbf{2. Tool and Format Fidelity}\\
If the environment targets a specific benchmark (e.g., tau2-bench, ToolSandbox):
\begin{itemize}
    \item Use the exact same tool schemas (function names, parameter names, return types)
\item Error messages and response formats should be consistent within the environment
\item The system prompt should set up the skill context clearly, but not replicate the exact benchmark's prompt
\item Data (database contents, user profiles, etc.) should be procedurally generated

The goal is to train the SKILL (e.g., multi-step reasoning, clarification, tool chaining), not to overfit to a particular benchmark's surface format.
\end{itemize}
This eliminates distribution shift between training and evaluation.

\vspace{0.5em}
\noindent \textbf{4. Procedural Generation}\\
All scenarios MUST be procedurally generated from a seed:
\begin{itemize}
    \item \texttt{reset(seed)} must produce a deterministic scenario for that seed
    \item Different seeds must produce meaningfully different scenarios (not just parameter swaps)
    \item The scenario space should be large enough that memorization is impossible
    \item Include randomized: database contents, user requests, constraint combinations, number of distractors
\end{itemize}

\vspace{0.5em}
\noindent \textbf{5. Skill Isolation}\\
Each environment should primarily stress ONE skill. If you identify multiple skills:
\begin{itemize}
    \item Create separate scenario generators per skill
    \item Assign probability weights (e.g., 40\% skill\_a, 35\% skill\_b, 25\% skill\_c)
    \item Reward functions can differ per skill, but the interface is the same
\end{itemize}

\vspace{0.5em}
\noindent \textbf{6. Multi-Skill Environments}\\
When combining multiple skills in one environment (like toolsandbox\_multiturn):
\begin{itemize}
    \item Sample skill per scenario at reset time based on probability weights
    \item Each skill gets its own scenario generators and reward logic
    \item Track which skill each scenario targets (for analysis)
    \item Aim for 1-2 skills max per environment --- beyond that, create separate envs
\end{itemize}

\vspace{1em}
\noindent \textbf{\large Deliverables}
\begin{enumerate}
    \item \textbf{\{game\_name\}\_game.py}: The environment file implementing \texttt{GameEnv}
    \item \textbf{Registration in game\_registry.py}: \newline \texttt{GameSpec} entry with proper \texttt{extract\_action}, \texttt{system\_prompt}, etc.
    \item \textbf{Reward verification}: Run 100+ rollouts using the vLLM server and report:
    \begin{itemize}
        \item Mean reward (target: 0.3-0.6 for base model)
        \item Reward standard deviation (target: $>$0.2)
        \item Per-skill breakdown if multi-skill
        \item \% of groups with reward variance (target: $>$60\%)
    \end{itemize}
\end{enumerate}

\vspace{1em}
\noindent \textbf{\large Verification Steps}

\noindent Before finalizing, verify using the vLLM server at \{VLLM\_URL\}:
\begin{lstlisting}[language=Python]
# 1. Run 100 rollouts and check reward distribution
rewards = []
for seed in range(100):
    env = YourGameEnv()
    env.reset(seed)
    # ... play episode with vLLM ...
    rewards.append(env.rewards[0])

print(f"Mean: {np.mean(rewards):.3f}")
print(f"Std:  {np.std(rewards):.3f}")
print(f"Distribution: {Counter([round(r, 1) for r in rewards])}")

# 2. Check GRPO compatibility (group variance)
informative_groups = 0
for group_seed in range(20):
    group_rewards = []
    for _ in range(8):  # group_size
        env = YourGameEnv()
        env.reset(group_seed)
        # ... play with temperature=1.0 ...
        group_rewards.append(env.rewards[0])
    if len(set(round(r, 2) for r in group_rewards)) > 1:
        informative_groups += 1

print(f"Informative groups: {informative_groups}/20 ({informative_groups/20*100:.0f}%)")
\end{lstlisting}

\vspace{1em}
\noindent \textbf{\large Reference Files}

\noindent Read these files to understand the architecture:
\begin{itemize}
    \item \texttt{game\_registry.py} --- \texttt{GameEnv} protocol, \texttt{GameSpec}, registration
    \item \texttt{train\_grpo\_optimized.py} --- How the training loop interfaces with games
\end{itemize}

\end{promptbox}

\section{Metrics Detail}
\label{app:metrics}

The overall pass rate for \tbench is defined as: 
\[
\mathrm{PassRate}_{\mathrm{overall}}
=
\frac{S_{\mathrm{Airline}} + S_{\mathrm{Retail}}}{N_{\mathrm{Airline}} + N_{\mathrm{Retail}}},
\]
where $S_{\mathrm{Airline}}$ and $S_{\mathrm{Retail}}$ denote the numbers of solved tasks in the two domains, and $N_{\mathrm{Airline}}$ and $N_{\mathrm{Retail}}$ denote the corresponding numbers of benchmark tasks.

The Pass@1 for \swe is defined as:
\[
\mathrm{Pass@1}
=
\frac{1}{N_{\mathrm{SWE}}}
\sum_{i=1}^{N_{\mathrm{SWE}}} \mathbf{1}\!\left[\text{patch}_i \text{ resolves issue } i\right],
\]
where the indicator is $1$ if and only if the single patch generated for instance $i$ applies cleanly and passes all of that instance's held-out tests, and $N_{\mathrm{SWE}}=500$ is the number of \swe instances.

The mean similarity and the perfect score for \tsb is defined as:
\begin{align}
\mathrm{MeanSimilarity}
&=
\frac{1}{N_{\mathrm{TS}}}
\sum_{i=1}^{N_{\mathrm{TS}}} \mathrm{Sim}_i, \\
\mathrm{PerfectRate}
&=
\frac{1}{N_{\mathrm{TS}}}
\sum_{i=1}^{N_{\mathrm{TS}}} \mathbf{1}\!\left[\mathrm{Sim}_i = 1\right].
\end{align}
where $\mathrm{Sim}_i$ denotes the final ToolSandbox similarity score for trajectory $\tau_i$, and $N_{\mathrm{TS}}$ is the number of evaluated ToolSandbox scenarios.

\section{Results with Stochastic Decoding}
\label{app:stochastic_decoding}

The main paper uses greedy decoding (temperature $T=0$, seed $42$) for deterministic reproducibility, which yields a single sample per task. We re-ran all baselines and \tool{} at temperature $0.7$ with $3$ seeds and report the mean $\pm$ 95\% confidence interval (CI) in Tables~\ref{tab:tau2_temp_ci} and~\ref{tab:tsb_temp_ci}. As shown in the tables, \tool{} consistently outperforms all baselines on both benchmarks, with non-overlapping CIs.

\begin{table}[t]
\small
\centering
\caption{Pass rate on \tbench{} under stochastic decoding (temperature $0.7$, $3$ seeds). We report mean $\pm$ 95\% CI.}
\label{tab:tau2_temp_ci}
\setlength{\tabcolsep}{1.5pt}
\begin{tabular}{@{}lccc@{}}
\toprule
\textbf{Approach} & \textbf{Airline (\%)} & \textbf{Retail (\%)} & \textbf{Overall (\%)} \\
\midrule
Base & $21.3 \pm 2.9$ & $36.3 \pm 2.5$ & $32.1 \pm 1.8$ \\
GRPO & $30.7 \pm 5.7$ & $38.9 \pm 3.3$ & $36.4 \pm 3.2$ \\
GEPA & $36.7 \pm 2.9$ & $38.6 \pm 2.2$ & $38.0 \pm 1.7$ \\
\midrule
\textbf{TRACE} & $\mathbf{44.0 \pm 5.0}$ & $\mathbf{47.1 \pm 2.5}$ & $\mathbf{46.1 \pm 3.2}$ \\
\bottomrule
\end{tabular}
\end{table}

\begin{table}[t]
\small
\centering
\caption{Pass rate and mean similarity on \tsb{} under stochastic decoding (temperature $0.7$, $3$ seeds). We report mean $\pm$ 95\% CI.}
\label{tab:tsb_temp_ci}
\begin{tabular}{lcc}
\toprule
\textbf{Model} & \textbf{Pass Rate (\%)} & \textbf{Mean Sim.} \\
\midrule
Base & $14.2 \pm 2.2$ & $0.40 \pm 0.03$ \\
GRPO & $16.3 \pm 1.9$ & $0.51 \pm 0.02$ \\
GEPA & $16.8 \pm 1.1$ & $0.52 \pm 0.01$ \\
\midrule
\textbf{TRACE} & $\mathbf{21.2 \pm 2.2}$ & $\mathbf{0.56 \pm 0.02}$ \\
\bottomrule
\end{tabular}
\end{table}

\section{Cross-Environment Transfer}
\label{app:cross_env_transfer}

We furthermore evaluate cross-environment transfer in Table~\ref{tab:cross_env_transfer}. A model trained with \tool{} or GRPO on \tbench{} is evaluated on a second tool-calling benchmark, \tsb{}, without re-training. The \tool{}-trained agent still transfers more effectively than the GRPO-trained one ($+1$ perfect score and $+0.052$ mean similarity), indicating that \tool{}'s targeted training yields capabilities that carry over to a new environment rather than merely overfitting to the training benchmark.

\begin{table}[t]
\small
\centering
\caption{Cross-environment transfer evaluated on \tsb{}. Agents trained with GRPO or \tool{} on \tbench{} are evaluated on \tsb{} without re-training; the bottom block reports in-domain \tsb{} training for reference. We report perfect score (similarity $=1.0$) out of $129$ scenarios and mean similarity.}
\label{tab:cross_env_transfer}
\begin{tabular}{llcc}
\toprule
\textbf{Method} & \textbf{Trained on} & \textbf{Perfect} & \textbf{Mean Sim.} \\
\midrule
Base & --- & 19/129 & 0.411 \\
\midrule
GRPO & \tbench & 19/129 & 0.430 \\
\textbf{TRACE} & \tbench & \textbf{20/129} & \textbf{0.482} \\
\midrule
GRPO & \tsb & 22/129 & 0.519 \\
\textbf{TRACE} & \tsb & \textbf{26/129} & \textbf{0.552} \\
\bottomrule
\end{tabular}
\end{table}

\section{Extended Capability Analysis}
\label{sec:appendix_capability}

While the analysis agent consistently selects the top five capabilities (as discussed in Section~\ref{sec:env_analysis}), it frequently discards competing categories. Candidates such as conditional reasoning, numerical reasoning, early termination, and information communication appeared much less frequently across the 10 independent runs. This long-tail distribution confirms that the target benchmark's failure modes are concentrated rather than uniform. 

Additionally, we note that the coverage counts reported in Figure~\ref{fig:capability_analysis}(b) are not mutually exclusive. A single failed benchmark trajectory often involves multiple missing capabilities---for example, an agent might fail to verify a precondition before executing a multi-step task. The contrastive analysis successfully disentangles these overlapping failures, allowing the framework to synthesize isolated micro-environments for each distinct deficit.

\section{Capability Details}
\label{app:cap_det}

\subsection{Training Capabilities and Trajectory Examples}

We identify four primary capability gaps through contrastive analysis of failed baseline trajectories on \tbench{}. Each capability targets a distinct failure mode.

\vspace{1mm}
\noindent\textbf{C1: Structured Data Reasoning.} The agent fails to parse or cross-reference structured records returned by tools. \textit{Example:} A user requests an economy flight departing after 11 AM. The search tool returns flights with nested price arrays per cabin class. The base model misreads which price corresponds to economy, computes the wrong total, and the booking fails with repeated payment errors.

\vspace{1mm}
\noindent\textbf{C2: Tool Calling Precision.} The agent identifies the correct tool but passes wrong arguments. \textit{Example:} A user requests a refund to their original payment method (credit card). The agent retrieves the order (showing \texttt{credit\_card\_3892} as payment) and the user profile (containing both \texttt{credit\_card\_3892} and \texttt{gift\_card\_1234}). When calling \texttt{return\_delivered\_order\_items}, it passes \texttt{gift\_card\_1234} instead of the correct \texttt{credit\_card\_3892}.

\vspace{1mm}
\noindent\textbf{C3: Multi-Step Task Completion.} The agent completes the first sub-task of a compound request then stops. \textit{Example:} A user asks to cancel two reservations and modify a third. The agent successfully cancels the first reservation, generates a closing statement ("If you need any further assistance\dots"), and never attempts the remaining two operations---entering a sycophancy loop with the user simulator until timeout.

\vspace{1mm}
\noindent\textbf{C4: Precondition Verification.} The agent executes state-changing actions without checking policy eligibility. \textit{Example:} A user requests cancellation of a basic economy flight booked 14 days ago with no insurance. The policy requires at least one of: booked within 24 hours, airline-cancelled flight, business class, or covered insurance. None apply, but the agent calls \texttt{cancel\_reservation} without checking any condition. The API does not enforce policy---the agent must independently verify eligibility.

\subsection{ToolSandbox Training Capabilities and Trajectory Examples}

We identify two primary capability gaps through analysis of base model trajectories on ToolSandbox:

\vspace{1mm}
\noindent\textbf{C1: Permission Error Recovery.} When a tool call returns a \texttt{PermissionError} due to a device setting conflict, the base model stops and reports the error to the user instead of resolving the underlying issue. \textit{Example:} The user asks ``Turn on wifi.'' The agent calls \texttt{set\_wifi\_status(on=True)}, which returns \texttt{PermissionError: Wifi cannot be turned on in low battery mode}. The base model halts---the conversation ends with only 2 messages. The correct behavior is to diagnose the blocker by calling \texttt{get\_low\_battery\_mode\_status} (returns \texttt{True}), disable it with \texttt{set\_low\_battery\_mode\_status(on=False)}, retry the original \texttt{set\_wifi\_status(on=True)}, and communicate success to the user.

\vspace{1mm}
\noindent\textbf{C2: Datetime Reasoning.} The base model skips the \texttt{timestamp\_to\_datetime\_info} tool and instead attempts to mentally decode Unix timestamps, consistently hallucinating the wrong date. \textit{Example 1:} The user asks ``Remind me to buy chocolate milk tomorrow 5PM.'' The agent calls \texttt{get\_current\_timestamp} (returns \texttt{1774511873}), then directly calls \texttt{datetime\_info\_to\_timestamp(year=2026, month=3, day=25, \dots)}---guessing the current date from the raw timestamp and getting the day wrong (March 25 instead of March 26), setting the reminder in the past. The correct behavior is to first call \texttt{timestamp\_to\_datetime\_info(1774511873)} which returns the exact date \texttt{\{year: 2026, month: 3, day: 26, \dots\}}, then compute tomorrow as March 27. \textit{Example 2:} The user asks ``How many days till Christmas Day?'' The base model calls \texttt{search\_holiday("Christmas Day", year=2023)}---defaulting to a training-data year---and computes -823 days (past). The correct behavior is to first determine the current year via \texttt{get\_current\_timestamp} followed by \texttt{timestamp\_to\_datetime\_info}, then search for the holiday in the correct year (2026), yielding 273 days.

\section{Synthetic Environment Example}
\label{app:synth_env_examples}

\subsection{Structured Data Reasoning (\texorpdfstring{$\tau^2$-Bench}{tau-squared-Bench})}
\definecolor{codegray}{rgb}{0.5,0.5,0.5}
\definecolor{backcolour}{rgb}{0.98,0.98,0.98}

\begin{center}
\begin{minipage}{0.95\textwidth}
\begin{tcolorbox}[
    colback=backcolour, 
    colframe=black, 
    title=Algorithm: Structured Data Reasoning Game (tau2-bench Aligned),
    fonttitle=\bfseries
]
\begin{lstlisting}[
    language=Python,
    basicstyle=\scriptsize\ttfamily, % Scriptsize is key for vertical fit
    breaklines=true,
    keywordstyle=\color{blue},
    commentstyle=\color{codegray},
    morekeywords={FUNCTION, IF, ELSE, RETURN, CLASS, METHOD, AND, OR, IN},
    showstringspaces=false,
    xleftmargin=5pt,
    xrightmargin=5pt
]
# =====================================================================
# 1. SCENARIO & DATABASE GENERATION
# =====================================================================

FUNCTION GenerateSDRScenario(seed, domain_choice):
    Initialize RNG(seed)
    domain = domain_choice OR RandomChoice(["Airline", "Retail"])
    
    # Select task type based on empirical failure weights
    task_type = WeightedChoice(SCENARIO_WEIGHTS[domain])
    
    # SYNTHETIC DB SYNTHESIS
    IF domain == "Airline":
        db = CreateFlightsDB()          # JSON: numbers, prices, seats
        user = CreateUser(reservations) # JSON: status, payments
    ELSE IF domain == "Retail":
        db = CreateProductsDB()        # JSON: 10+ variants, stock
        user = CreateUser(orders)      # JSON: history, tracking

    # CONSTRUCT GOAL & CONSTRAINTS
    # Gap 3: Conditional Fallback | Gap 6: Cross-Entity Match 
    # Gap 7: Progressive Disclosure 
    
    RETURN SDRScenario(initial_msg, user_sys_prompt, db, expected_tool, expected_answer)


# =====================================================================
# 2. MULTI-TURN ENVIRONMENT LOOP
# =====================================================================

CLASS StructuredDataGame:
    METHOD Reset(seed):
        scenario = GenerateSDRScenario(seed)
        env_db = Copy(scenario.db)
        user_sim = InitializeLLMUser(scenario.user_sys_prompt)
        RETURN scenario.initial_message

    METHOD Step(agent_action):
        IF agent_action IS "tool_call":
            result = ExecuteTool(agent_action.name, agent_action.args, env_db)
            RETURN tool_output(result)
            
        ELSE IF agent_action IS "respond_to_user":
            user_response = user_sim.GenerateResponse(agent_action.text)
            
            IF "###STOP###" IN user_response:
                TerminateGame(EvaluateReward())
            ELSE:
                RETURN user_response


# =====================================================================
# 3. REWARD & VALIDATION LOGIC
# =====================================================================

FUNCTION EvaluateReward(conversation, scenario, final_db):
    # STEP 1: VERIFY COMMUNICATION (RELIABLE REPORTING)
    comm_pass = All(val IN agent_messages FOR val IN scenario.communicate_info)

    # STEP 2: VERIFY STATE MUTATION (DATABASE INTEGRITY)
    IF scenario.expects_mutation:
        # Prevents rewards for "hallucinated" success messages
        db_match = (Hash(final_db) == Hash(ReplayGoldAction(scenario.db)))
        
        IF db_match AND comm_pass: RETURN 1.0  # Success
        IF db_match:               RETURN 0.3  # Correct action, bad info
        RETURN 0.0
    
    # STEP 3: INFORMATION RETRIEVAL TASKS
    RETURN 1.0 IF comm_pass ELSE 0.0
\end{lstlisting}
\end{tcolorbox}
\end{minipage}
\end{center}

\subsection{Error Recovery (\tsb)}

\definecolor{codegray}{rgb}{0.5,0.5,0.5}
\definecolor{backcolour}{rgb}{0.98,0.98,0.98}

\begin{center}
\begin{minipage}{0.95\textwidth}
\begin{tcolorbox}[
    colback=backcolour, 
    colframe=black, 
    title=Algorithm: TEC Game v2 (Tool-use \& Error Recovery),
    fonttitle=\bfseries
]
\begin{lstlisting}[
    language=Python,
    basicstyle=\scriptsize\ttfamily,
    breaklines=true,
    keywordstyle=\color{blue},
    commentstyle=\color{codegray},
    morekeywords={FUNCTION, IF, ELSE, RETURN, CLASS, METHOD, AND, OR, IN, TRY, EXCEPT},
    showstringspaces=false
]
# =====================================================================
# 1. SCENARIO GENERATION & SKILL TARGETS
# =====================================================================

FUNCTION GenerateTECScenario(seed):
    Initialize RNG(seed)
    roll = RandomValue(0, 1)
    
    # Skill A: Tool-then-Communicate (40%)
    IF roll < 0.40:
        RETURN Scenario(skill="communicate", verify_keywords=[...])
        
    # Skill B: Error Recovery Chain (40%)
    ELSE IF roll < 0.80:
        # PermissionError triggered by system settings (e.g. Low Battery)
        RETURN Scenario(skill="recovery", blocker="low_battery")
        
    # Combined: Recovery + Communication (20%)
    ELSE:
        RETURN Scenario(skill="combined")

# =====================================================================
# 2. MULTI-TURN ENVIRONMENT LOOP
# =====================================================================

CLASS TECGame:
    METHOD Step(agent_action):
        IF agent_action IS "tool_call":
            # Execute tool against synthetic ToolSandbox DB
            result = ExecuteTool(agent_action.name, agent_action.args)
            
            # If low_battery is True, certain tools return PermissionError
            # Agent must diagnose and call 'set_low_battery_mode_status'
            RETURN tool_result_to_agent(result)
            
        ELSE:
            # Final text response to user
            reward = EvaluateTEC(agent_action.text, self.db, self.scenario)
            TerminateGame(reward)

# =====================================================================
# 3. REWARD EVALUATION
# =====================================================================

FUNCTION EvaluateTEC(agent_response, final_db, scenario):
    # ACTION SCORE: Was the blocked service successfully enabled?
    IF scenario.skill IN ("recovery", "combined"):
        action_score = 1.0 IF TargetServiceIsActive(final_db) ELSE 0.0
    ELSE:
        action_score = 1.0 IF ToolWasCalled(scenario.target) ELSE 0.0
            
    # COMM SCORE: Did the agent provide the result to the user?
    # Matches keywords from tool output (e.g. phone numbers, coordinates)
    comm_score = KeywordMatchRatio(agent_response, scenario.keywords)
    
    # Weighted Reward: Combines execution and reporting
    RETURN (0.6 * action_score) + (0.4 * comm_score)
\end{lstlisting}
\end{tcolorbox}
\end{minipage}
\end{center}

\end{document}